\documentclass[final]{cvpr}

\usepackage{times}
\usepackage{epsfig}
\usepackage{graphicx}
\usepackage{amsmath}
\usepackage{amssymb}

\usepackage{comment}
\usepackage{color}

\usepackage{makecell}
\usepackage{textcomp}
\usepackage{array,multirow,graphicx}
\usepackage{float}
\usepackage{xcolor}

\usepackage{subcaption}

\usepackage[pagebackref=true,breaklinks=true,colorlinks,bookmarks=false]{hyperref}

\begin{document}

\definecolor{revcolor}{RGB}{255,50,0}
\definecolor{rev2color}{RGB}{0,50,255}
\newcommand\rev[1] {\emph{\textcolor{revcolor}{#1}}}
\newcommand\REV[1] {\emph{\textcolor{rev2color}{#1}}}
\newcommand{\miss}{\textcolor{red}{REF}}
\newcommand{\due}[1]{{\sethlcolor{red}\hl{#1}}}
\newcommand{\unsure}[1]{{\textcolor{cyan}{\textit{#1} (unsure)}}}
\newcommand{\parahead}[1]{\textbf{#1}:\ }
\newcommand{\tabhead}[1]{\par\textbf{#1}}
\newcommand{\TODO}[1]{{\textcolor{red}{[TODO: #1]}}}
\newcommand{\change}[1]{{\textcolor{purple}{#1}}}
\newcommand{\ignore}[1]{{}}

\makeatletter
\def\thanks#1{\protected@xdef\@thanks{\@thanks
        \protect\footnotetext{#1}}}
\makeatother

\title{Learning Speech-driven 3D Conversational Gestures from Video}

\author{Ikhsanul Habibie\textsuperscript{1}, Weipeng Xu\textsuperscript{2}, Dushyant Mehta\textsuperscript{1}, Lingjie Liu\textsuperscript{1}, Hans-Peter Seidel\textsuperscript{1}, \\
Gerard Pons-Moll\textsuperscript{1}, Mohamed Elgharib\textsuperscript{1}, Christian Theobalt\textsuperscript{1}\\
\textsuperscript{1}Max Planck Institute for Informatics, Saarland Informatics Campus, Germany\\
\textsuperscript{2}Facebook Reality Labs, USA \\

\thanks{This work was supported by the ERC Consolidator Grant 4DRepLy (770784). Gerard Pons-Moll is funded by the Deutsche Forschungsgemeinschaft (DFG. German Research Foundation) - 409792180.}
}

\maketitle

\begin{abstract}
We propose the first approach to automatically and jointly synthesize both the synchronous 3D conversational body and hand gestures, as well as 3D face and head animations, 
of a virtual character from speech input. Our algorithm uses a CNN architecture that leverages the inherent correlation between facial expression and hand gestures. 
Synthesis of conversational body gestures is a multi-modal problem since many similar gestures can plausibly accompany the same input speech. To synthesize plausible body gestures in this setting, we train a Generative Adversarial Network (GAN) based model that measures the plausibility of the generated sequences of 3D body motion when paired with the input audio features. We also contribute a new way to create a large corpus of more than 33 hours of annotated body, hand, and face data from in-the-wild videos of talking people. To this end, we apply state-of-the-art monocular approaches for 3D body and hand pose estimation as well as dense 3D face performance capture to the video corpus. In this way, we can train on orders of magnitude more data than previous algorithms that resort to complex in-studio motion capture solutions, and thereby train more expressive synthesis algorithms. Our experiments and user study show the state-of-the-art quality of our speech-synthesized full 3D character animations.  
\end{abstract}

\begin{figure*}
    \centering
    \includegraphics[width=1\linewidth]{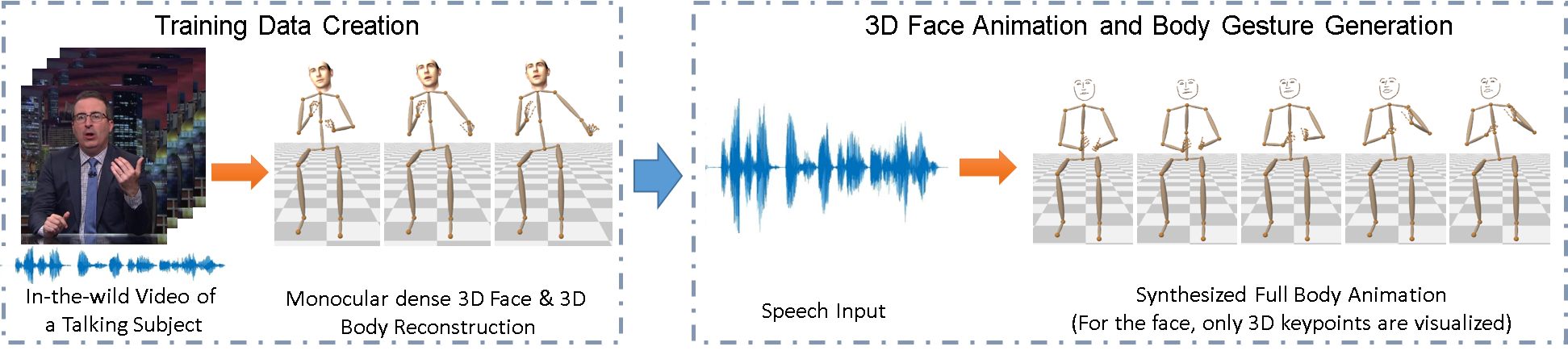}
    \caption{We propose the first approach to jointly synthesize the synchronous 3D conversational body gestures and 3D face animations of a virtual character from speech input. It is trained using our new contributed 3D facial expression, body, and hand pose annotation for a large corpus of in-the-wild video of talking people.}
   \label{fig:teaser}
    \vspace{-0.5cm}
\end{figure*}

\section{Introduction}
\label{sec:intro}

Virtual human characters are a crucial component in many computer graphics applications, such as games or shared virtual environments. Traditionally, their generation requires a combination of complex motion capture recordings and tedious work by animation experts to generate plausible appearance and movement.
The particular challenges include the animation of the conversational body gestures of a talking avatar, as well as the facial expressions that accompany the audio in conveying the emotion and mannerisms of the speaker.
Both are traditionally achieved by manually specified key-frame animation. 
Automated tools for animating facial expressions and body gestures directly from speech would drastically ease the effort required, and allow non-experts to author higher quality character animations. Further on, such tools would enable users to drive real-time embodied conversational virtual avatars of themselves populating shared virtual spaces, and animate them with on-the-fly facial expressions and body gestures in tune with speech. 
In psycho-linguistics studies, it has been shown that user interfaces showing avatars with plausible facial expressions, body gestures, and speech are perceived as more believable and trustworthy~\cite{van1998persona}. 
It was also shown that non-verbal behavior is important for conveying information~\cite{goldin1999role}, providing a view into the speaker's internal state, and both speech and body gestures are tightly correlated, arising from the same internal process~\cite{mcneill_2000,kendon_2004}. 

Prior work on speech-driven virtual characters has been limited either to the generation of co-verbal body gestures through heuristic rule-based~\cite{Marsella2013virtual} or learning-based \cite{Ferstl2018investigating,Levine2009real,Levine2010GestureC} approaches, or the generation of facial expressions~\cite{Karras2017} and head movements~\cite{Sadoughi2017meaningful} in tune with speech. Many learning-based approaches use motion and gesture training data captured in a studio with complex motion capture systems~\cite{lee2019handmotiondataset,Levine2009real,Levine2010GestureC,takeuchi2017creating,Ferstl2018investigating, ferstl2019multi, alexanderson2020style}. In this way, it is hard to record large corpora of data reflecting gesture variation across subjects, or subject-specific idiosyncrasies revealing only in long term observation.   

We propose the first approach to jointly generate synchronized conversational 3D gestures of the arms, torso, and hands, as well as a simple but expressive 3D face and head movement of an animated character from speech. It is based on the 
following contributions: 
{\bf (1)} We contribute a new way to capture annotated 3D training data
from more than 33 hours of in-the-wild videos of talking subjects, which was used for learning a purely 2D gesturing model, without face expression synthesis, before~\cite{ginosar2019gestures}. To create ground truth, we apply monocular in-the-wild 3D body pose reconstruction~\cite{mehta2019xnect}, 3D hand pose reconstruction~\cite{zhou2019monocular}, and monocular dense 3D face reconstruction~\cite{garrido2016reconstruction} on these videos. 
These annotations will be made available.  
{\bf (2)} We introduce a new CNN architecture that has a common encoder for facial expression, body, and hands gesture which learns the inherent correlation between them and three decoder heads to jointly generate realistic motion sequences for face, body, and hands.
In addition to facial expressions and head poses in tune with audio, it synthesizes plausible conversational gestures, such as beat gestures that humans use to emphasize spoken words, and gestures that reflect mood and personal conversational style. Note that, like in previous works, the goal is not to generate gestures relating to semantic speech content, 
or carrying specific language meaning, like in sign language. 
{\bf (3)} Synthesis of body gestures is a multi-modal problem; several gestures could accompany the same utterance. To prevent convergence to the mean pose in training and ensure expressive gesture synthesis, the prior 2D work of Ginosar~\etal~\cite{ginosar2019gestures} used adversarial training \cite{Goodfellow2014generative}.
We improve upon this work by not only designing a discriminator that can measure whether the synthesized body and hand gestures look natural, but also the plausibility of the synthesized gestures when paired with the ground truth audio features.
We evaluate our approach through extensive user study, where the participants rate our results as more natural and in-tune than the baseline methods.

\section{Related Work}
\label{sec:related}

Prior work looked at the problems of body gesture and face animation synthesis from audio input largely in separation. Problem settings differ, as conversational gesture synthesis is a much more multi-modal setting than speech-driven face animation where the viseme to phoneme mapping is much more unique. 
In this paper, we look at them in combination.

\paragraph{Speech-Driven Body Gestures and Head Motion} 

The prior art can be grouped into rule-based and data-driven methods.
The seminal work by Cassell \etal ~\cite{cassell1994animated} and Cassell \cite{cassell2000embodied} show that automatic body gesture and facial expression generation of a virtual character can be synchronized with the audio by using a set of manually defined rules. 
Other work incorporate linguistic analysis \cite{cassell2004beat} into an extendable rule-based framework.
Marsella \etal \cite{Marsella2013virtual} develop a rule-based system to generate body gesture (and facial expression) by analyzing the content of the text input and audio. 
However, such methods heavily rely on the study of language-specific rules and cannot easily handle non-phoneme sounds.

To overcome these problems, data-driven approaches, which do not rely on expert knowledge in the linguistic domain, have attracted increasing attention.
Neff \etal~\cite{Neff2008Gesture} propose a method to create a person-specific gesture script using manually annotated video corpora, given the spoken text and performer's gesture profile. The gesture script is then used to animate a virtual avatar.
Levine \etal~ \cite{Levine2009real} use a complex motion capture setup to capture 45 minutes of training data and trained a Hidden Markov Model to select the most probable body gesture clip based on the speech prosody in real time. 
Later, Levine \etal~\cite{Levine2010GestureC} argue that the prosody does not carry enough information to determine the exact gesture form. Instead, speech generally encodes kinematic gesture features, such as velocity and spatial extent. 
Based on this assumption, they first mapped the audio signal into a latent kinematic feature space using a variant of Hidden Conditional Random Fields (CRF). The learned model is then used to select a gesture sequence via reinforcement learning approach.
Similarly, Mariooryad and Busso \cite{mariooryad2012generating} use a combination of Dynamic Bayesian Networks (DBN) to synthesize head pose and eyebrow motion from speech. Sadoughi \etal~\cite{Sadoughi2014speech} extended this approach by modeling discourse functions as additional constraints of the DBNs.
Bayesian Networks was also used by Sadoughi and Busso \cite{Sadoughi2017speech} to generate body gestures by a combination of speech signal, semantic discourse functions, and prototypical motions.
Bozkurt \etal~\cite{Bozkurt2013multimodal} use Hidden Semi-Markov Models that can learn speech-to-gesture relation asynchronously. 
Sadoughi \etal~\cite{Sadoughi2017meaningful} use a learning-based approach that can leverage text-to-speech (TOS) system to synthesize head motion and propose a method that can solve the mismatch between real and synthetic speech during training. 
Chiu and Marsella \cite{Chiu2011how} train a Conditional Restricted Boltzmann Machine (CRBM) to directly synthesize sequences of body poses from speech. Chiu and Marsella later proposed using
Gaussian Process Latent Variable Models (GPLVM) to learn a low dimensional embedding to select the most probable body gestures from a given speech input \cite{Chiu2014gesture}. 

In recent years, deep learning has demonstrated its superiority in automatically learning discriminative features from big data.
Bidirectional LSTM was used by Takeuchi \etal \cite{takeuchi2017speech}, Hasegawa \etal \cite{hasegawa2018evaluation}, and Ferstl and McDonnell \cite{Ferstl2018investigating} to synthesize body gestures from speech. Similarly, Haag and Shimodaira \cite{Haag2016bidirectional} used LSTM to synthesize head motion from speech. 
A GAN-based approach to synthesize head movement of a talking agent was proposed by Sadoughi and Busso \cite{Sadoughi2018novel}. 
Kucherenko \etal \cite{kucherenko2019analyzing} propose a denoising autoencoder to learn lower dimensional representation of body motion and then combines it with an audio encoder to perform audio-to-gesture synthesis at test time. 
Lee \etal~\cite{lee2019handmotiondataset} contribute a large scale motion capture dataset of synchronized body-finger motion and audio, and propose a method to predict finger motion based on both audio and arm position as input. Ferstl \etal~\cite{ferstl2019multi} use a multi-objective adversarial model and make use of a classifier that is trained to predict the gesture phase of the motion to improve gesture synthesis quality. 

Recent work also try to incorporate text-based semantic information to improve generation quality of body gestures from speech \cite{Kucherenko2020gesticulator, Yoon2020Trimodal}. Alexanderson \etal~\cite{alexanderson2020style} propose a normalizing flow-based generative model that can synthesize 3D body gesture from speech input. Their system can generate multiple plausible gesture from the same audio and allows some degrees of control to their synthesis, such as the speed and symmetry of the gesture. Ahuja \etal \cite{ahuja2020style} show that a single learning-based mixture model can be trained to perform gesture style transfer between multiple speakers. In contrast, our focus is to find the best solution of predicting all relevant body modalities from audio using a single framework, which is a challenging problem even when trained in a person-specific manner.

Deep learning approaches typically require a large scale training corpus of audio and 3D motion pairs, which is usually captured with complex and expensive in-studio motion capture systems.
To tackle this, Ginosar~\etal~\cite{ginosar2019gestures} propose a learning-based speech-driven generation of 2D upper body and hand gesture model from a large scale in-the-wild video collection.
With this solution, they are able to build an order of magnitude larger corpus from community video. 
Similarly, the method of Yoon~\etal~\cite{Yoon2018robots} was trained using ground truth 2D poses extracted from TED Talk videos via OpenPose \cite{cao2017realtime}.  
Their model employs a Bidirectional LSTM to map audio input into a sequence of 2D human body pose.
In addition, they used a lifting approach to derive 3D motion constraints for the synthesized character. 
In our work, we contribute additional 3D face, hand, and body annotation for the dataset of Ginosar \etal~\cite{ginosar2019gestures}.
Furthermore, in contrast to existing methods, we synthesize not only the 3D upper body and hand gestures, but also head rotation and facial expression of the speaker.

\paragraph{Speech-Driven Facial Expressions}

Current techniques can be classified into: 1) face model-based \cite{Liu15,Pham18,Cha2018TFM,Taylor2017,Tzirakis19,VOCA2019} and 2) non-model based. Model-based approaches parameterize expressions in terms of blendshapes, and estimate these parameters from the audio input. Non-model based approaches, however, directly map the audio into 3D vertices of a face mesh \cite{Karras2017} or 2D point positions of the mouth \cite{suwajanakorn2017obama}. In Karras \etal~\cite{Karras2017}, an LSTM is used to learn this mapping, and in Suwajanakorn \etal~\cite{suwajanakorn2017obama}, final photorelistic results are generated. 
Cudeiro \etal~\cite{VOCA2019} use DeepSpeech voice recognition \cite{deepspeech} to produce an intermediate representation of the audio signal. This is then regressed into the parameters of the FLAME face model \cite{flame}. Taylor \etal~\cite{Taylor2017} use an off-the-shelf speech recognition method to map the audio into phoneme transcripts. A network is trained to translate the phonemes into the parameters of a reference face model. Tzikrakis \etal~\cite{Tzirakis19} use a Deep Canonical Attentional Warping (DCAW) to translate the audio into expression blendshapes. Pham \etal~\cite{Pham18} directly maps the audio to the blendshape parameters even though their results suffer from strong jitter. While current audio-driven facial expression techniques produce interesting results, most of them show results on voice data recorded in controlled studios with minimal background noise \cite{Liu15,Pham18,Karras2017,Cha2018TFM,Taylor2017,VOCA2019}. Cudeiro~\etal~\cite{VOCA2019} showed interesting results in handling different noise levels. Nevertheless, there is currently no audio-driven technique that estimates high quality facial expressions in-the-wild, as well as estimating the head motion and body conversational gestures. 
Our method uses the face model as the first category.
In contrast to other methods, we apply a simple but effective approach to jointly learn the 3D head and face animation with body gestures, by directly regressing the facial parameters captured from a large corpus of community video.

\section{Dataset Creation}
\subsection{Creating 3D Annotations from Video}
\label{sec:data}

\begin{figure}[t]
    \centering
    \includegraphics[width=1\linewidth]{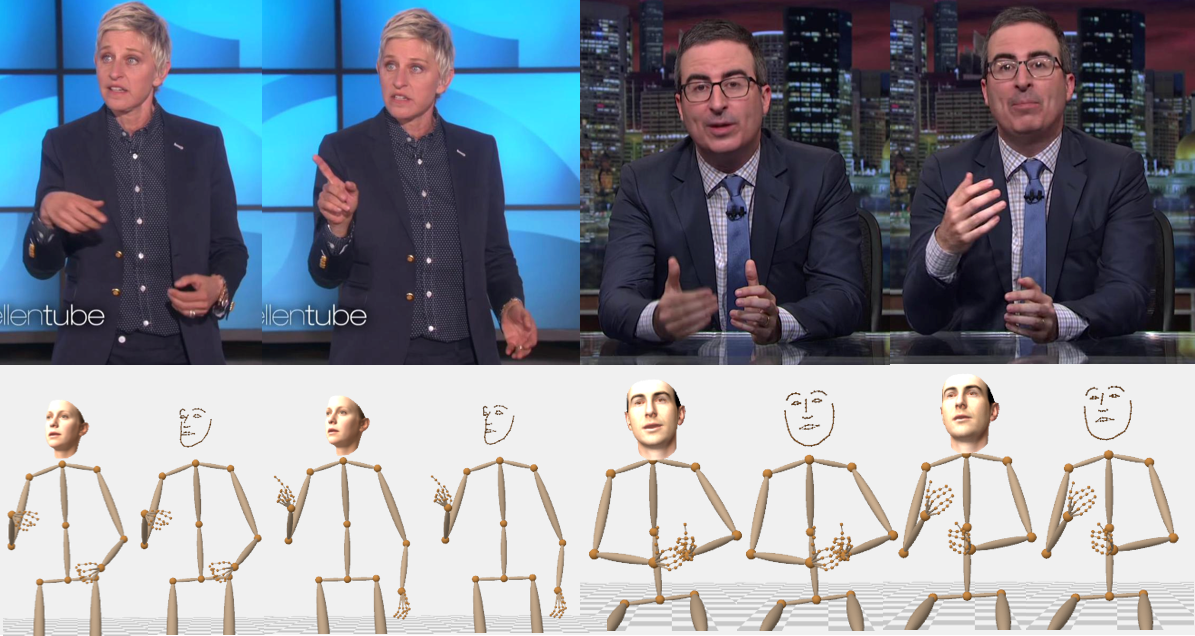}
    \caption{Training data. We annotate the in-the-wild conversational video corpus of Ginosar \etal~\cite{ginosar2019gestures} with 3D parameters of dense facial reconstruction, hand poses, and body poses.}
    \label{fig:dataset}
    \vspace{-0.5cm}
\end{figure}

A major bottleneck for previous speech-driven animation synthesis work is the generation of sufficient training data. Many methods resort to complex in-studio capture of face and full-body motion with multi-camera motion capture systems. We therefore propose the first approach to extract automatic annotations of 3D face animation parameters, 3D head pose, 3D hands, and 3D upper body gestures from a large corpus of community video with audio. In this way, much larger training corpora spanning over long temporal windows and diverse subjects can be created more easily. 

In particular, we use the dataset of Ginosar \etal \cite{ginosar2019gestures} which features 144 hours of in-the-wild video of 10 subjects (e.g. talk show hosts) talking into the camera in both standing and sitting poses. From these videos, Ginosar \etal extracted 2D keypoints of the arms and hands, as well as 2D sparse face landmarks. They used a subset of these annotations to train a network synthesizing only 2D arm and finger motion from speech. While showing the potential of speech-driven animation, their approach does not synthesize 3D body motion; does not synthesize 3D motions of the torso, such as leaning, which is an element of personal speaker style; and does not predict 3D head pose and detailed face animation parameters. To train a method jointly synthesizing the latter more complete 3D animation parameters in tune with input speech, we annotate the dataset with state-of-the-art 3D face performance capture and monocular 3D body and hand pose estimation algorithms, see Fig.~\ref{fig:dataset}. 

For monocular dense 3D face performance capture, we use the optimization-based tracker \cite{garrido2016reconstruction} that predicts parameters of a parametric face model, specifically: $64$ expression blend shape coefficients, $80$ PCA coefficients of identity geometry, $80$ PCA coefficients of face albedo, $27$ incident illumination parameters, and $6$ coefficients for 3D head rotation and position.
The face tracker expects tightly cropped face bounding boxes as input. We use the face tracker from Saragih \etal \cite{Saragih2011} for bounding box extraction and temporally filter the bouding box locations; we experimentally found this to be more stable than using the default 2D face landmarks in Ginosar's dataset for bounding box tracking. 
For training our algorithm, we only use the face expression coefficients $\mathcal{\theta}_{Face} \in \mathbb{R}^{64}$ and head rotation 
coefficients $\mathbf{R}\in SO(3)$ (we use the 3D head position found by the body pose tracker).

For 3D body capture, we need an approach robust to body self-occlusions, occlusions by other people, occlusions of the body by a desk (sitting poses by talk show hosts) or occlusions by camera framing not showing the full body, even in standing poses. We therefore use the XNect \cite{mehta2019xnect} monocular 3D pose estimation approach designed to handle these cases. 
Specifically, in each video frame we extract 3D body keypoint predictions from \emph{Stage II} of XNect for the $13$ upper body joints (2 for head, 3 for each arm, 1 for neck, 1 for spine, 3 for hip/pelvis). This results in a 39-dimensional representation $\mathcal{K} \in \mathbb{R}^{39}$ for the body pose. We group the head rotation $\mathcal{R}$ predicted by the face tracker together with 3D the body keypoints $\mathcal{K}$ in a 42-dimensional vector $\mathcal{\theta}_{Body} \in \mathbb{R}^{42}$.

To perform hand tracking, we employ the state-of-the-art monocular 3D hand pose estimation method of Zhou \etal \cite{zhou2019monocular}. 
To ensure good prediction results, we first tightly crop the hand images using the 2D hand keypoint annotations provided by Ginosar \etal \cite{ginosar2019gestures} before feeding it to the 3D hand pose predictor. Since the hands can be occluded or out of view, we also employ an off-the-shelf cubic interpolation method to fill-in potentially missing 3D hand pose information as long as they lie between two annotated frames which are at most 8 frames apart. This results in 21 joints prediction for each hands, which we group into a 126-dimensional vector $\mathcal{\theta}_{Hand} \in \mathbb{R}^{126}$.

To improve the robustness of our data due to potential monocular tracking issues such as occlusions, we exclude the data if the prediction confidence of the face landmarks or hand keypoints within a certain number of frames falls below a given threshold. This is obtained by reinterpreting the maximum value of the 2D joint heatmap prediction of the body parts produced by the tracker as a confidence measure. We also remove 4 out of 10 subjects provided by Ginosar \etal \cite{ginosar2019gestures} due to the low resolution of the videos which lead to poor quality 3D dense face reconstruction results. Our final 3D dataset consists of more than 33 hours of videos from 6 subjects.

\subsection{Audio features pre-preprocessing}
\label{sec:audio_data}
Similar to Suwajanakorn \etal \cite{suwajanakorn2017obama}, we compute the MFC coefficients of each input video frame with an overlapping temporal sliding window after normalizing the audio using FFMPEG \cite{ffmpeg,ffmpeg_norm}. We make use of CMU Sphinx \cite{Lamere03thecmu} for computing the coefficients, and use 13 MFC coefficients and an additional feature to account for the log mean energy of the input. These, together with their temporal first derivatives, yield a 28-dimensional vector $\mathcal{F}_{MFC} \in \mathbb{R}^{28}$ representing the speech input at each time step.
MFCC encodes the characteristics of how human speech is perceived, which make it useful for a wide range of applications such as speech recognition. Encoding the characteristics of speech perception make MFC coefficients a good representation for predicting facial expressions because modulation of face shapes is a part of the speech production process. For predicting body gestures, the change of MFCC features over the sequence carries the rhythm information needed to produce beat gestures.

\section{Method}
\label{sec:main}

\begin{figure*}[t]
    \centering
    \includegraphics[width=0.82\textwidth]{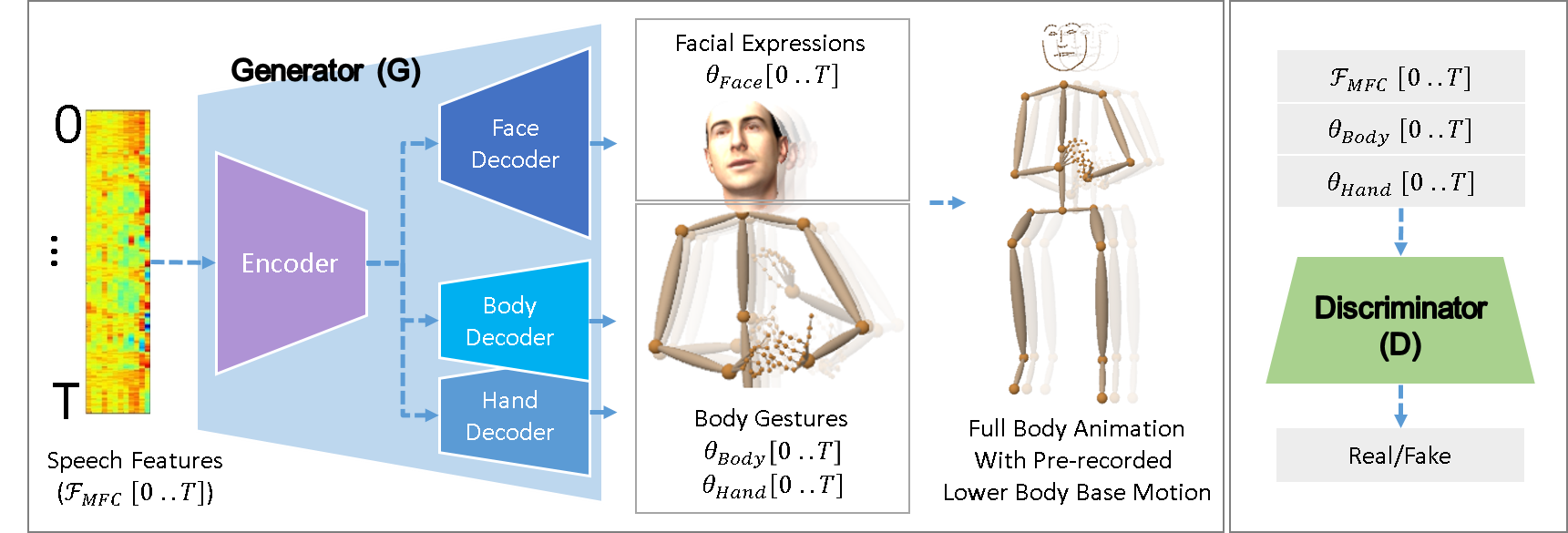}
    \caption{Our approach produces a temporal sequence of 3D facial expression parameters, head orientation, and 3D keypoints of the upper body and hands given a speech signal as input. We employ an adversarial loss in which the discriminator network tries to distinguish whether the input audio and body pose features are real or generated by the generator network.
    }
    \vspace{-0.5cm}
    \label{fig:pipeline}
\end{figure*}

Our approach produces a temporal sequence of 3D facial expression parameters, head orientation, 3D body, and 3D hand pose keypoints given a speech signal as input. Temporal variations in these aforementioned parameters contain the gestural information. 
As described in section ~\ref{sec:audio_data}, the speech input is pre-processed to yield MFC based feature frames $\mathcal{F}_{MFC}[t] \in \mathbb{R}^{28}$ for each discrete time step $t$. We indicate the facial expression parameters at each time step as $\mathcal{\theta}_{Face}[t] \in \mathbb{R}^{64}$, 3D keypoints for both hands as  $\mathcal{\theta}_{Hand}[t] \in \mathbb{R}^{126}$, and the head orientation and 3D body keypoints are represented together as $\mathcal{\theta}_{Body}[t] \in \mathbb{R}^{42}$.
The temporal sequences are sampled at $15 Hz$. 

\subsection{Network Architecture}
\label{subsec:arch}
Similar to other adversarial learning-based approaches, our model consists of 2 main neural networks that we refer as the generator network $G$ and discriminator network $D$.

We employ a 1D convolutional Encoder-Decoder architecture for the generator network $G$ to map the input audio feature sequence $\mathcal{F}_{MFC}[0:T]$ to 3D face expression parameter sequence $\mathcal{\theta}_{Face}[0:T]$, 3D body pose parameter sequence $\mathcal{\theta}_{Body}[0:T]$, and 3D hand pose parameter sequence $\mathcal{\theta}_{Hand}[0:T]$ which is also trained in a supervised manner. 

Our 1D convolutional architecture for the generator $G$ is adapted from a reference implementation~\cite{pytorch-unet} of the U-Net~\cite{ronneberger2015u} architecture originally proposed for 2D image segmentation.
Our architecture utilizes a single encoder, comprised of $8$ 1D [Conv-BN-ReLU] blocks with a kernel size of 3, and is interleaved with MaxPool after every second block except the last. The last block is followed by an upsampling layer (nearest neighbour).
Each of face, body, and hand sequences utilize a separate decoder. The decoders are symmetric with the encoder, and comprised of $7$ 1D [Conv-BN-ReLU] blocks and a final 1D convolution layer, interleaved with upsampling layers after every second block. The decoders, being symmetric with the encoder, utilize skip connectivity from the corresponding layers in the encoder.

The discriminator network is designed to predict whether its input audio and pose features are real or not. This network comprised of $6$ 1D [Conv-BN-ReLU] blocks with a kernel size of 3, and is interleaved with MaxPool after every second block. Afterwards, it is followed a linear and sigmoid activation layers.

Refer to the supplementary document for exact details of the architecture. A schema is shown in Figure~\ref{fig:pipeline}. 

\subsection{Training Details}
For each sequence of sampled speech features $\mathcal{F}_{MFC}[0~..~T-1]$, and annotated 3D face expression parameter sequence $\mathcal{\theta}_{Face}[0~..~T-1]$, 3D body pose parameter sequence $\mathcal{\theta}_{Body}[0~..~T-1]$, and 3D hand pose parameter sequence $\mathcal{\theta}_{Hand}[0~..~T-1]$, we extract 64 frame ($\approx 4$sec) sub-sequences in a sliding window manner, with a 1 to 5 frame overlap between consecutive sub-sequences depending on the number of data points of the subject.
Each mini-batch for training comprises of a random sampling of such 64-frame sub-sequences extracted from all training sequences. 
We use Adam for training, with a learning rate of $5e-4$, a mini-batch size of 25, and trained until 300,000 iterations per subject.
Since the generator network is fully convolutional, at deployment our network can handle input speech features of arbitrary duration.

\noindent
\parahead{Training Objective}
We supervise our generator network $G$ with the following loss terms:

\vspace{-0.5cm}

\begin{equation}
    \mathcal{L}_{Reg} = w_1*\mathcal{L}_{Face} + w_2*\mathcal{L}_{Body} + w_3*\mathcal{L}_{Hand}.
\end{equation}

\noindent
$\mathcal{L}_{Face}$ is the L2 error of facial expression parameters
$$
\mathcal{L}_{Face} = \sum_{t=0}^{T-1}\|\mathbf{\theta}_{Face}[t]-\hat{\mathbf{\theta}}_{Face}[t]\|_2.
$$
$\mathcal{L}_{Body}$ is the L1 error of 3D body keypoint locations and head orientation, and $\mathcal{L}_{Hand}$ is the L1 error of 3D hand keypoint locations
$$
\mathcal{L}_{Body} = \sum_{t=0}^{T-1}\|\mathbf{\theta}_{Body}[t]-\hat{\mathbf{\theta}}_{Body}[t]\|_1, 
$$

$$
\mathcal{L}_{Hand} = \sum_{t=0}^{T-1}\|\mathbf{\theta}_{Hand}[t]-\hat{\mathbf{\theta}}_{Hand}[t]\|_1.
$$

\noindent
We define $w_1 = 0.37$, $w_2 = 600$, and $w_3 = 840$ to ensure that each term is equally weighted during training.

In practice, we observe that only employing L1 or L2 error for body keypoints results in less expressive gestures, as has also been pointed out in prior work on 2D body gesture synthesis of Ginosar~\etal~\cite{ginosar2019gestures}.
Inspired by the adversarial training approach of Ginosar~\etal \cite{ginosar2019gestures}, we show that incorporating an adversarial loss using a discriminator network $D$ which is trained to judge whether an input pose is real or fakely generated by the generator $G$, can lead to more expressive gestures that are also in-sync with the speech input. When trained together with the generator network in a minimax game scenario, it will push the generator to produce a higher quality 3D body and hand pose synthesis in order to fool the discriminator. We follow similar approach to the work of Ferstl~\etal \cite{ferstl2019multi} by using not only the pose, but also the audio features as input to the discriminator. This way, the discriminator is not only tasked to measure if the input gesture looks real, but it also needs to determine if the gesture is in-sync with the input audio features or not. Since the multi-modality of the body gestures mainly occurs for the body and hands, we exclude the facial expression parameters from the adversarial loss formulation:

\vspace{-0.4cm}

\begin{multline}
\mathcal{L}_{Adv}(G, D) = \mathbb{E}_{\mathcal{F}_{MFC}}[\log (1 - D(\mathcal{F}_{MFC}, G^*(\mathcal{F}_{MFC})))] \\ 
 + \mathbb{E}_{\mathcal{F}_{MFC},\theta_{Body},\theta_{Hand}}[\log D(\mathcal{F}_{MFC},\theta_{Body},\theta_{Hand})] 
\end{multline}

\noindent
where $G^{*}$ indicates that we only use the predicted $\theta_{Body}$ and $\theta_{Hand}$ outputs of the original generator network $G$.

Combined with the direct supervision loss, our overall loss is:

\vspace{-0.4cm}

\begin{equation}
    \mathcal{L} = \mathcal{L}_{Reg} + w \cdot \min_{D} \max_{G} \mathcal{L}_{Adv}(G, D)
\end{equation}

\noindent
where $w$ is set to be 5.

Note that our networks are trained on subject specific training sets in order to capture the particular gesture characteristics of the subject.

\section{Results}
\label{sec:results}

Our proposed approach addresses essential aspects of animating virtual humans: synthesizing facial expressions, body, and hand gestures in tune with speech. For visualization of our results, as well as for the user study, to allow observers to focus on the face and body motion, we render an abstract 3D character that showcases all the important skeletal and facial elements without the risk of falling in the uncanny valley, following similar approaches in prior work~\cite{Levine2009real,ginosar2019gestures}. 
Since our approach only predicts upper body motion, we fuse it with a pre-recorded base motion of the lower body in both sitting and standing scenarios. 

\noindent
\parahead{Assessing the Generation Quality}
Since the synthesis of conversational gestures is a multi-modal problem, direct comparison with the tracked annotations would not be meaningful for all aspects of the synthesized results, particularly for evaluating the realism of the synthesized gestures.
We evaluate our method through extensive user studies to judge the quality and the plausibility of our results, and we compare it to various baselines.
Further, we measure the prediction of the facial expressions by comparing the 3D lip keypoints extracted from selected vertices of the predicted dense 3D face model against the automatically generated ground truth lip keypoints that we obtained from the source image. Please see the accompanying video for extensive audio-visual results.
 
\begin{table}[ht]
\small
    \caption{
    User study result measuring both the naturalness and synchronization between the synthesized face+body+hand gesture and speech.
}
	\label{table:user_study} 
	\centering
	\begin{tabular}{c|c|c}
		\hline
	    \textbf{Method}	& \makecell{Naturalness} & \makecell{Synchrony.} \\
		\hline
		\makecell{Ground truth} & $\mathbf{4.29 \pm 0.86}$ & $\mathbf{4.39 \pm 0.77}$ \\ \hline
	    \makecell{Conv.net. direct regress.} & $3.54 \pm 1.11$ & $3.78 \pm 1.08$ \\ \hline
	    \makecell{LSTM (adapted from \cite{shlizerman2018audio})} & $3.15 \pm 1.03$ & $3.21 \pm 1.11$ \\ \hline
	    \makecell{Adv. loss on \\ veloc. (adapted from \cite{ginosar2019gestures})} & $3.03 \pm 0.98$ & $3.38 \pm 0.95$ \\ \hline
	    \hline
	    \makecell{Adv. loss on \\ audio+3D pose (ours)}  & $\mathbf{4.05} \pm \mathbf{0.85}$ & $\mathbf{4.00} \pm \textbf{0.91}$ \\ \hline
	\end{tabular}
	    \vspace{-0.3cm}
\end{table} 

\subsection{Baseline Comparisons}

We evaluate our approach against other methods that perform body gesture prediction which use audio features as input. Other baseline methods are trained using the same MFCC features described in \ref{sec:audio_data}.
Our first baseline is the direct regression 1D CNN model of our proposed network architecture without using the adversarial loss.
Next, we compare our method against a Recurrent Neural Network (RNN)-based Long Short-term Memory (LSTM) architecture by Shlizerman \etal \cite{shlizerman2018audio} which was originally designed to temporally predict 2D hand and finger poses. Since the original method was not designed to handle multi-modal data, we decided to train three LSTM models for face, body, and hand gesture separately on our 3D data.

We trained an adaptation of Ginosar \etal \cite{ginosar2019gestures} using our proposed model and trained the adversarial loss to distinguish between the real and fake synthesis of the gesture in the velocity space similar to their proposed approach and use this version as our baseline comparison.
We also compare our method against the work of Alexanderson \etal \cite{alexanderson2020style} by retraining their method on our in-the-wild 3D data.
Their model was originally trained on clean mocap data of 3D body pose without face or hand annotations. 
We found that the model is sensitive to the hyperparameters used. Because of this, we decided to only train it on the body and hand data to simplify the problem. We manually searched for an optimal set of hyperparameters that can produce the best results in terms of naturalness and synchronization based on the recommendation of the authors. Following their instruction, we conducted multiple experiments by varying the number of units $H$ between $512$, $700$, and $800$ and the number of flow-steps $K$ from $8$ up to $16$. We found that the MoGlow-based model produces the best results when using the number of units $H=800$ and number of steps $K=10$. 

Please also refer to our supplementary video for the qualitative results of the baseline methods.

\subsection{User Study Evaluation on the Gesture Synthesis}
We conducted two separate user studies for the qualitative evaluation of our proposed method.

For the first user study, we compare methods that synthesize the 3D face, body, and hand gestures from audio. In this study, the participants were shown 3 out of 6 video sequences (12 seconds/sequence) synthesized by our proposed method, baselines, and the ground truth (tracked) annotations. This study involved 67 participants. Each user was asked to judge the naturalness and the synchronization between the audio and the generated 3D face and body gestures on a scale of 1 to 5, with 5 being the most plausible and 1 being the least plausible.

As shown in Table~\ref{table:user_study}, the ground truth sequences are perceived as both the most natural and in-tune with the input speech compared to other synthesized gesture videos, which is rated at $4.29 \pm 0.86$ and $4.39 \pm 0.77$, respectively. Compared to other baseline methods, the participants agree that our results look more natural and in tune with the speech audio with the score of $4.05 \pm 0.85$ in term of naturalness and $4.00 \pm 0.91$ in terms of synchronization with the speech.

\begin{table}[ht]
\small
    \caption{
    User study result measuring both the naturalness and synchronization between the synthesized body+hand gesture and speech. The users were specifically asked to ignore the quality of the facial expression.
}
    \vspace{-0.1cm}
	\label{table:user_study2} 
	\centering
	\begin{tabular}{c|c|c}
		\hline
	    \textbf{Method}	& \makecell{Naturalness} & \makecell{Synchrony.} \\
		\hline

	    \makecell{MoGlow \cite{alexanderson2020style})} & $2.88 \pm 1.02$ & $3.11 \pm 1.13$ \\ \hline
	    \makecell{Ours}  & $\mathbf{4.01} \pm \mathbf{0.82}$ & $\mathbf{3.93} \pm \textbf{0.92}$ \\ \hline
	\end{tabular}
	    \vspace{-0.3cm}
\end{table} 

We also conducted a second user study evaluating only the synthesis of the 3D body and hand gestures and compare our method with the MoGlow-based model of Alexanderson \etal \cite{alexanderson2020style}. For this study, the participants were specifically asked to ignore the quality of the face expressions in the video. To ensure a fair comparison, all videos presented in this study were synthesized by using the 3D facial expression predicted by our method. Similar to the first study, each of the 45 participants were asked to rate the quality of the gestures from 3 out of 6 possible videos for each method on a scale between 1 to 5. As shown in Table \ref{table:user_study2}, our method is rated as both more natural and in-sync with the audio.

\begin{figure*}[ht]
    \centering
    \includegraphics[width=0.8\linewidth]{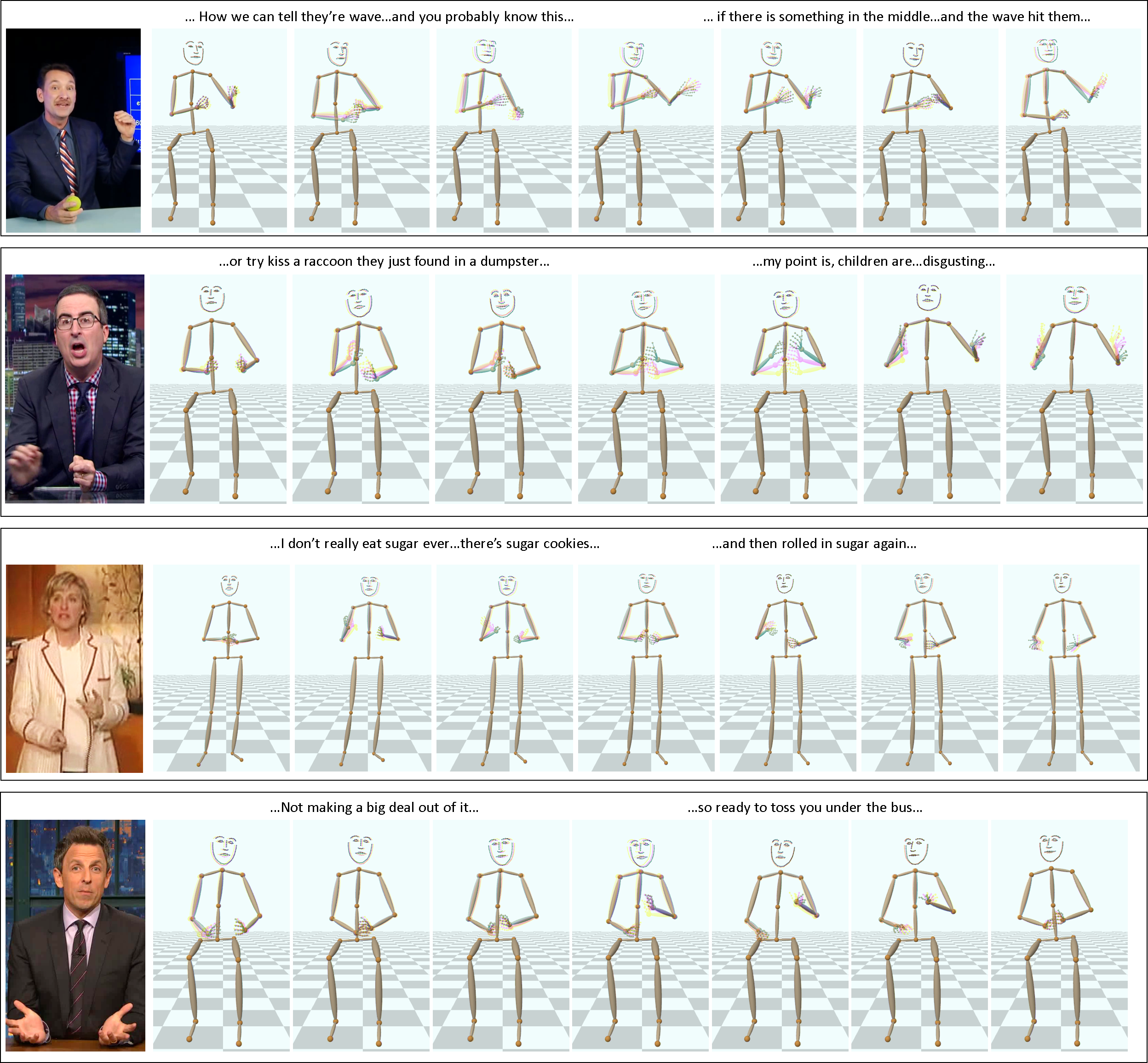}
    \caption{Qualitative results of our approach. Note that our approach is subject specific, and separate trained models are used for each subject for which representative sequence is shown. Motion visualization is based on the Harris shutter effect.
    }
    \label{fig:output_sequence}
    \vspace{-0.1cm}
\end{figure*}

\begin{table*}[ht]
    \caption{
    Quantitative comparison to baseline methods for lip motion prediction error against the ground truth (in mm).}
	\label{table:ablation} 
	\centering
	\begin{tabular}{c|c|c|c|c|c|c}
		\hline
	    \textbf{Method}	& \makecell{Oliver} & \makecell{Meyers}   & \makecell{Ellen} & \makecell{Kubinec}  & \makecell{Stewart} & O'Brien \\
		\hline
	    Conv. network direct regression & \textbf{0.29} & \textbf{0.35} & 0.29 & \textbf{0.28} & \textbf{0.39} & \textbf{0.37} \\ \hline
	    LSTM (adapted from \cite{shlizerman2018audio}) & 0.3 & 0.36 & 0.30 & 0.32 & 0.41 & 0.39 \\ \hline
	    Adv. loss on velocity (adapted from 
	    \cite{ginosar2019gestures}) & \textbf{0.29} & \textbf{0.35} & 0.3 & 0.29 & 0.39 & 0.38 \\ \hline
	    Random & 0.49 & 0.57 & 0.47 & 0.43 & 0.57 & 0.52 \\ \hline
	    \hline
	    Adv. loss on audio+3D pose (ours) & \textbf{0.29} & \textbf{0.35} & \textbf{0.28} & \textbf{0.28} & \textbf{0.39} & \textbf{0.37} \\ \hline
	\end{tabular}
	    \vspace{-0.3cm}
\end{table*}

\subsection{Facial Expression Evaluation}

In Table~\ref{table:ablation}, we compare the 3D lip keypoints of the generated face vertices corresponding to the facial expressions predicted by various approaches against the image-based face tracker's 3D lip keypoints in a neutral head pose. The comparison was performed on the whole test set which consists of 578 sequences (12 seconds/sequence) across all subjects. As a sanity check baseline, we also compute the difference between the optimization-based tracked annotations of one sequence, to the optimization-based annotations on a \emph{different} sequence \emph{chosen randomly}. The evaluation shows that our proposed method achieves similar or slightly better performance against other proposed baselines.

This result also demonstrate that our unified whole-body architecture is suitable for simultaneous face expression synthesis at decent quality, and better than simultaneous face synthesis with other body gesture synthesis architectures. Note here that we are not claiming that our design advances the state-of-the-art in face-only expression synthesis. This is outside the scope of our work and left for future work.

\section{Discussion}
\label{sec:discussion}

Although mouth expressions are strongly correlated with speech, the rest of the intended generation targets such as body gestures do not have a one-to-one mapping. Coupled with the noisy nature of our monocular data, as observed in our experiments, this multi-modal nature of the problem makes both designing and analyzing a stable expressive model challenging. 
We also observe that a lower value of $L1$ or $L2$ loss on the validation set does not always guarantee to produce a qualitatively better gesture synthesis, which further shows the importance of the adversarial loss.
The choice of hyperparameters such as the design of encoder-decoder architecture and the weighting of the loss terms can also result to different gesture synthesis characteristics.
For example, asymmetrical update between the generator and discriminator with different number of iterations can  lead to different but equally plausible gesture synthesis.

Following the success of the proposed adversarial loss, we further argue that the discriminator network can also be used independently as a plausibility metric to rate the quality of a gesture synthesis from speech, similar to how the inception score is used \cite{salimans2016improved}. Such a method can be trained with the objective to classify in-sync and off-sync audio gesture pairs with various kinds of body gesture noise and plausibility characteristics. If we can create such a dataset, this model can be used to quantitatively evaluate other proposed audio-to-gesture synthesis models in the future.

\section{Conclusion}
\label{sec:conclusion}

We propose the first approach for full 3D face, body, and hand gesture prediction from speech to automatically drive a virtual character or an embodied conversational agent. We leverage monocular dense face reconstruction and body pose reconstruction approaches on in-the-wild footage of talking subjects to acquire training data for our learning-based approach, generating 3D face, body, and hand pose annotations for $\approx33$ hours of footage.
Our key insight on incorporating an adversarial penalty not only on the 3D pose but also its combination with the audio input allows us to successfully generate expressive body gestures that are in-sync with the speech.
\newline

\appendix

\noindent
\textbf{\Large Supplementary Material}

\setcounter{equation}{0}
\setcounter{figure}{0}
\renewcommand{\thefigure}{S\arabic{figure}}
\setcounter{table}{0}
\renewcommand{\thetable}{S\arabic{table}}
\setcounter{section}{0}

\section{On 3D Annotations of In-The-Wild Video}
\begin{figure*}[ht]
\centering
    
\begin{subfigure}{.40\textwidth}
  \centering
  \includegraphics[width=\linewidth]{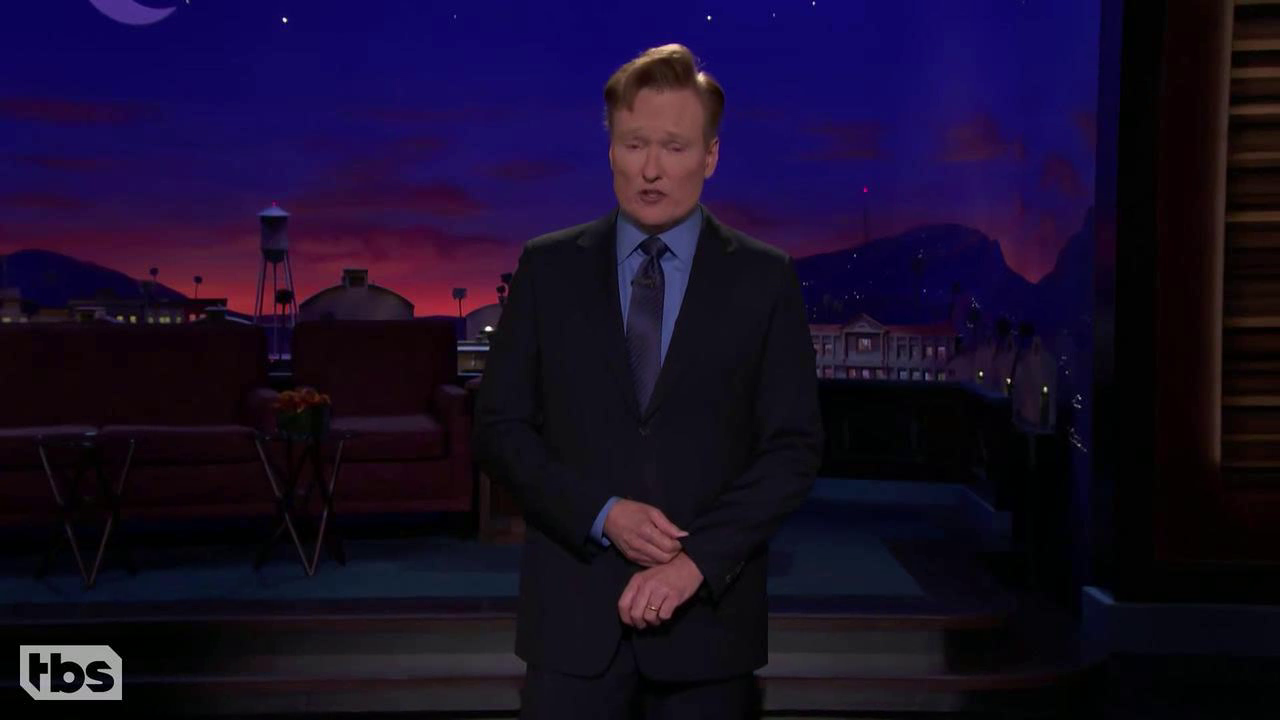}
\end{subfigure}
\begin{subfigure}{.34\textwidth}
  \centering
  \includegraphics[width=\linewidth]{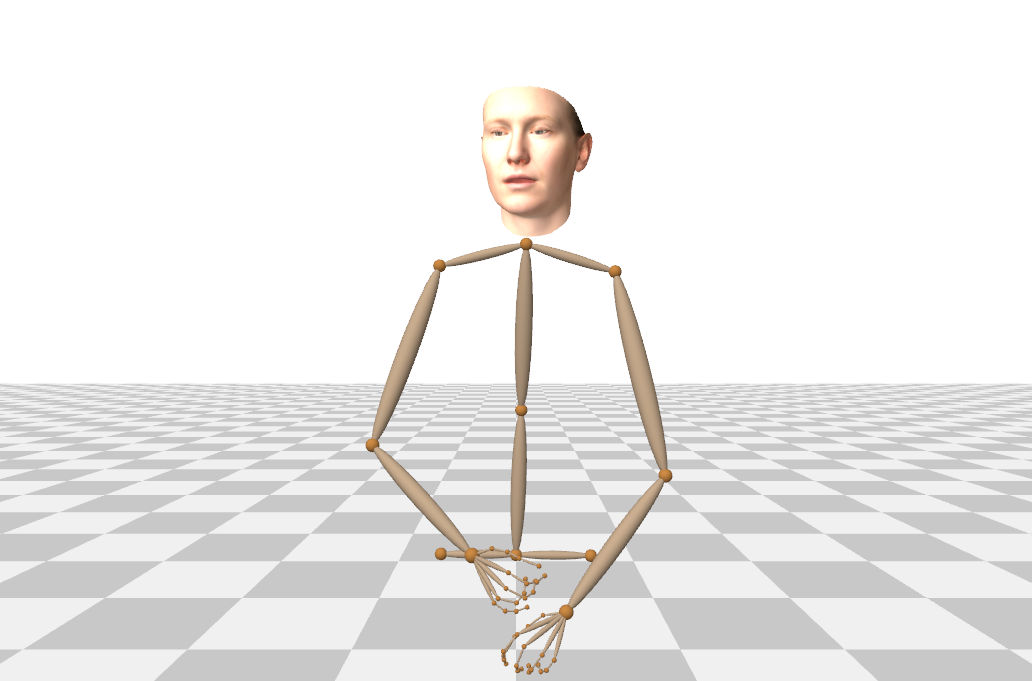}
\end{subfigure}

\vspace{1mm}

\begin{subfigure}{.40\textwidth}
  \centering
  \includegraphics[width=\linewidth]{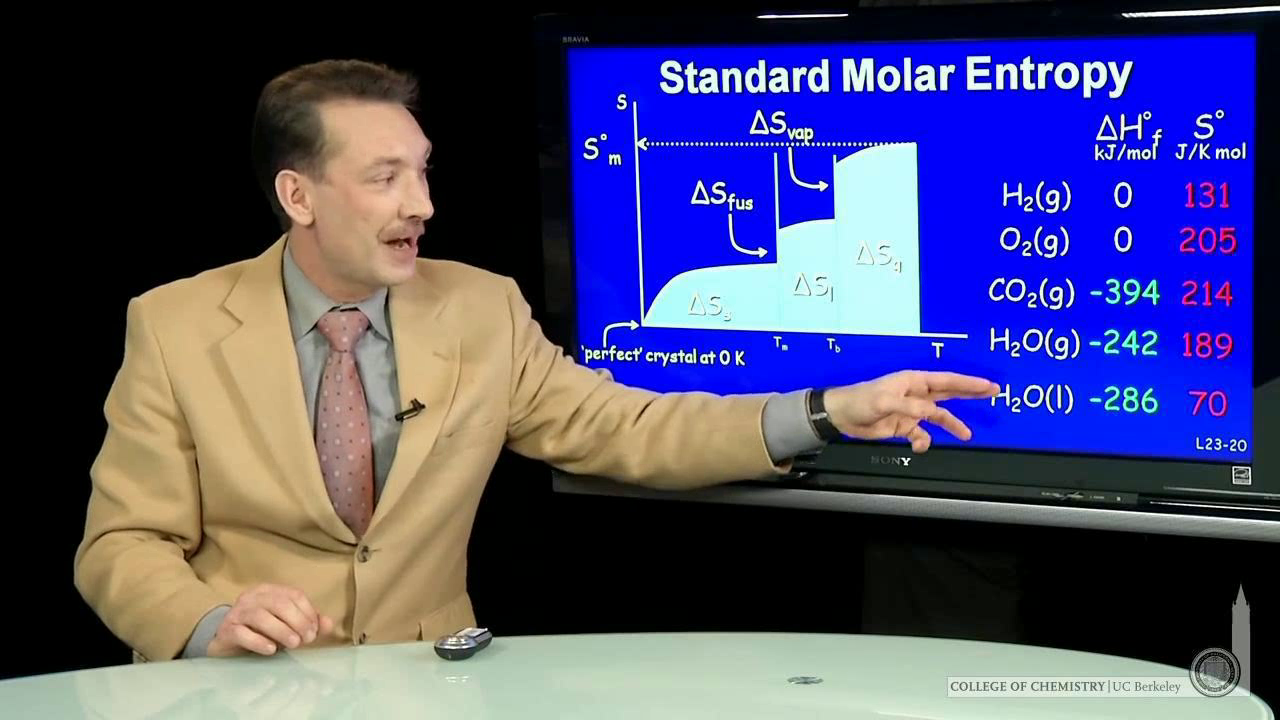}
\end{subfigure}
\begin{subfigure}{.34\textwidth}
  \centering
  \includegraphics[width=\linewidth]{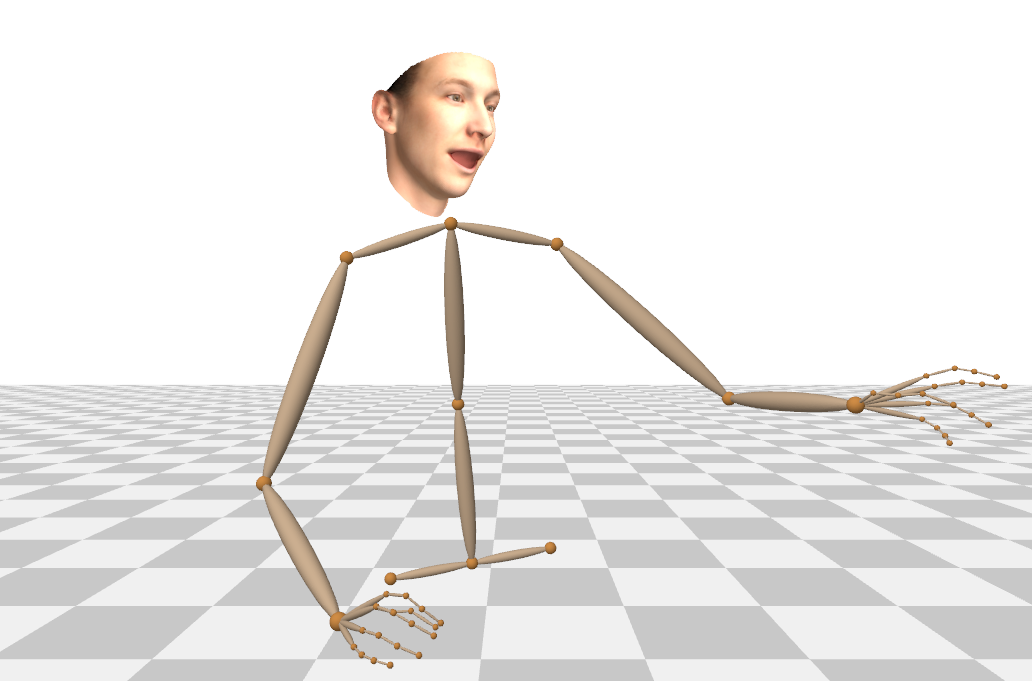}
\end{subfigure}

\vspace{1mm}

\begin{subfigure}{.40\textwidth}
  \centering
  \includegraphics[width=\linewidth]{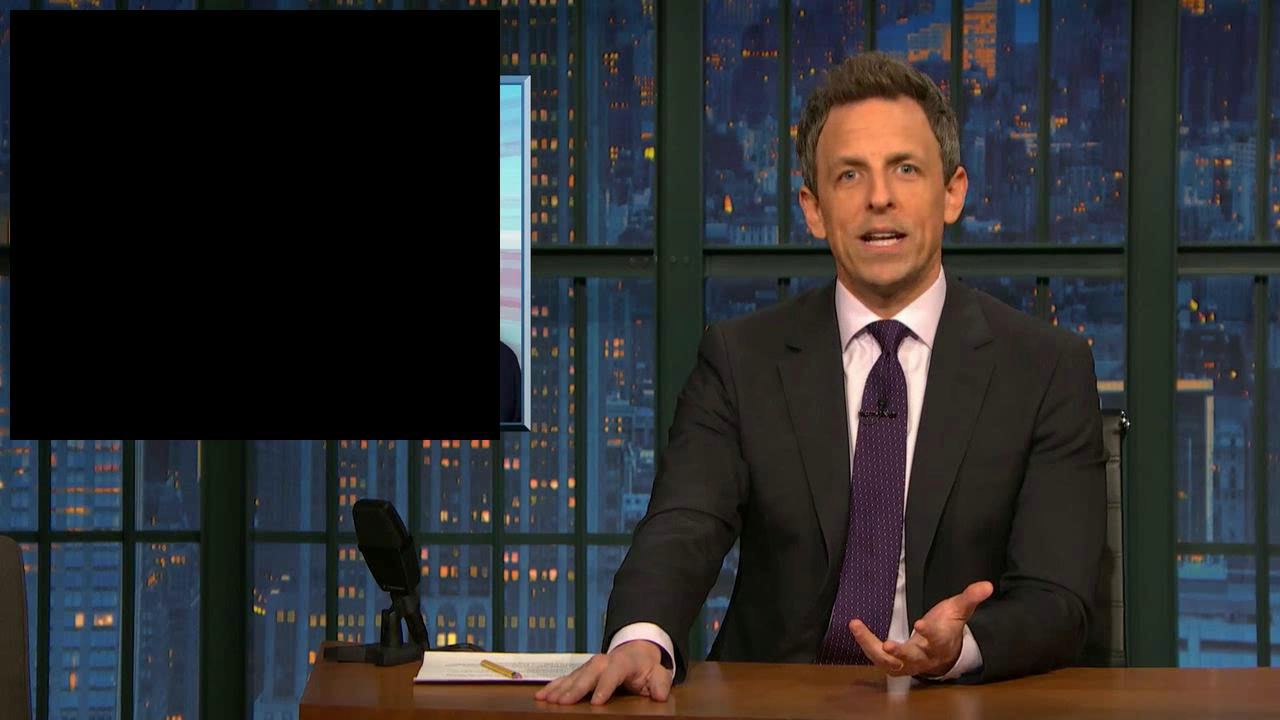}
\end{subfigure}
\begin{subfigure}{.34\textwidth}
  \centering
  \includegraphics[width=\linewidth]{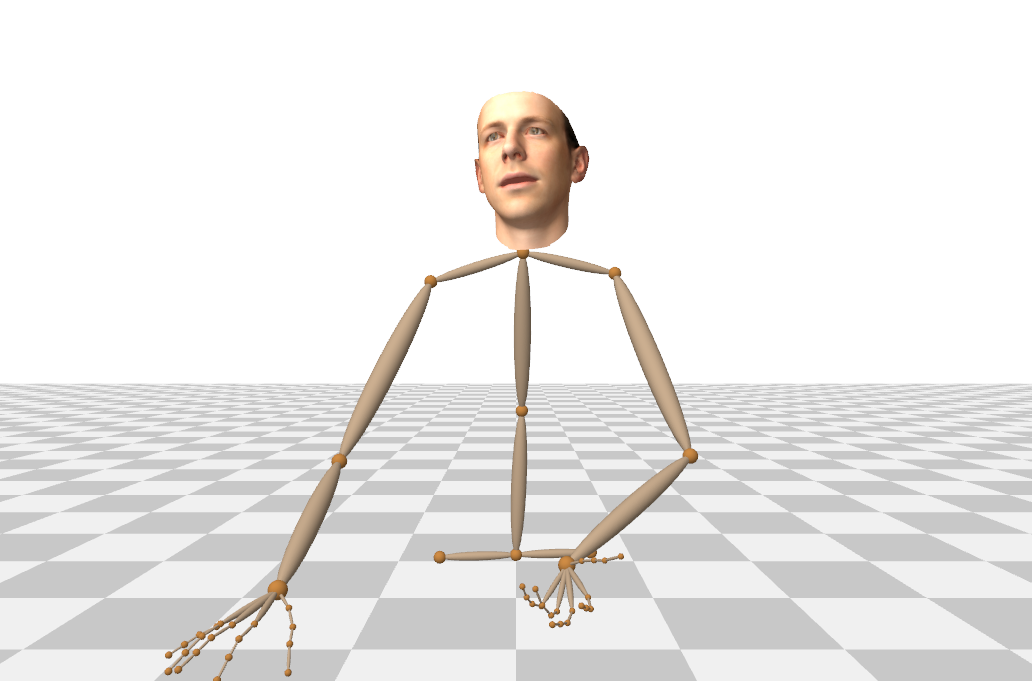}
\end{subfigure}

\vspace{1mm}

\begin{subfigure}{.40\textwidth}
  \centering
  \includegraphics[width=\linewidth]{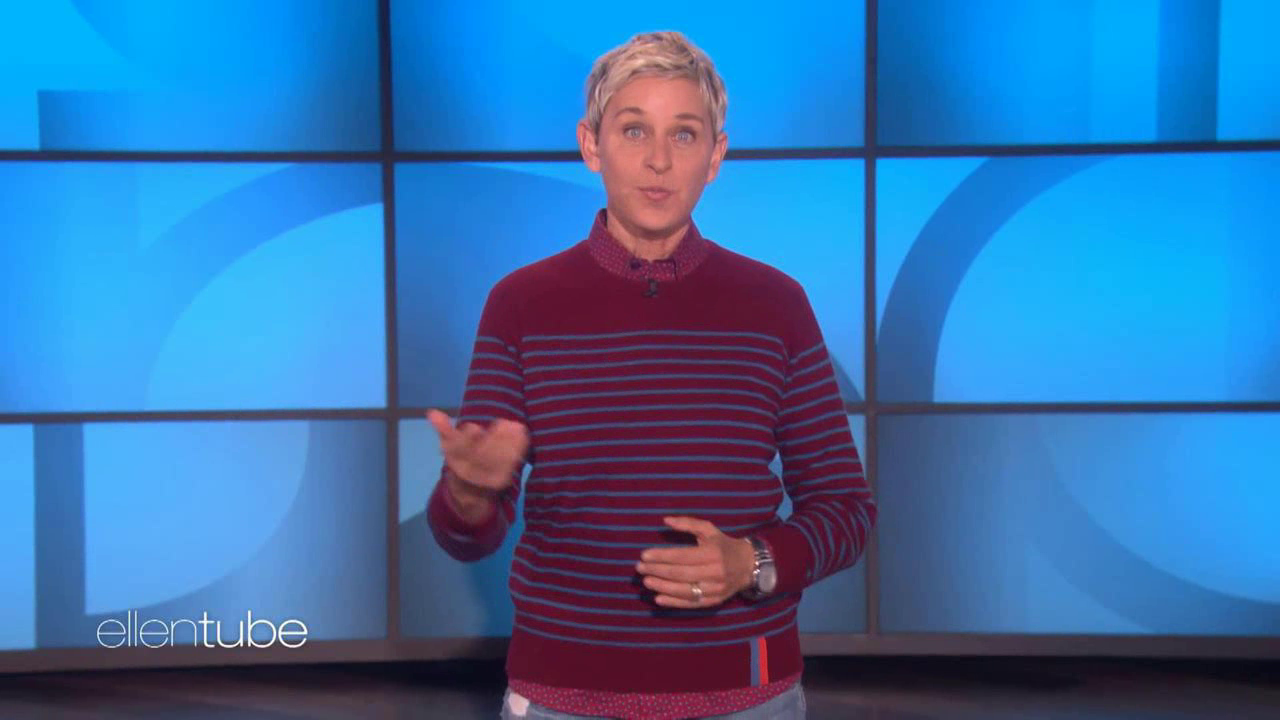}
\end{subfigure}
\begin{subfigure}{.34\textwidth}
  \centering
  \includegraphics[width=\linewidth]{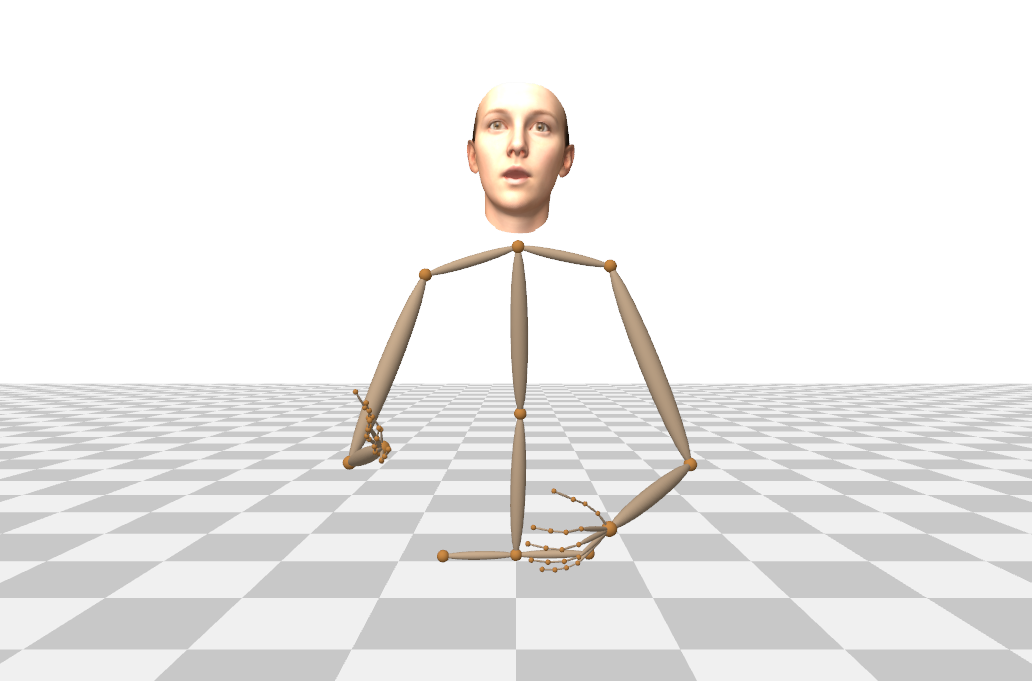}
\end{subfigure}

\vspace{1mm}

\begin{subfigure}{.40\textwidth}
  \centering
  \includegraphics[width=\linewidth]{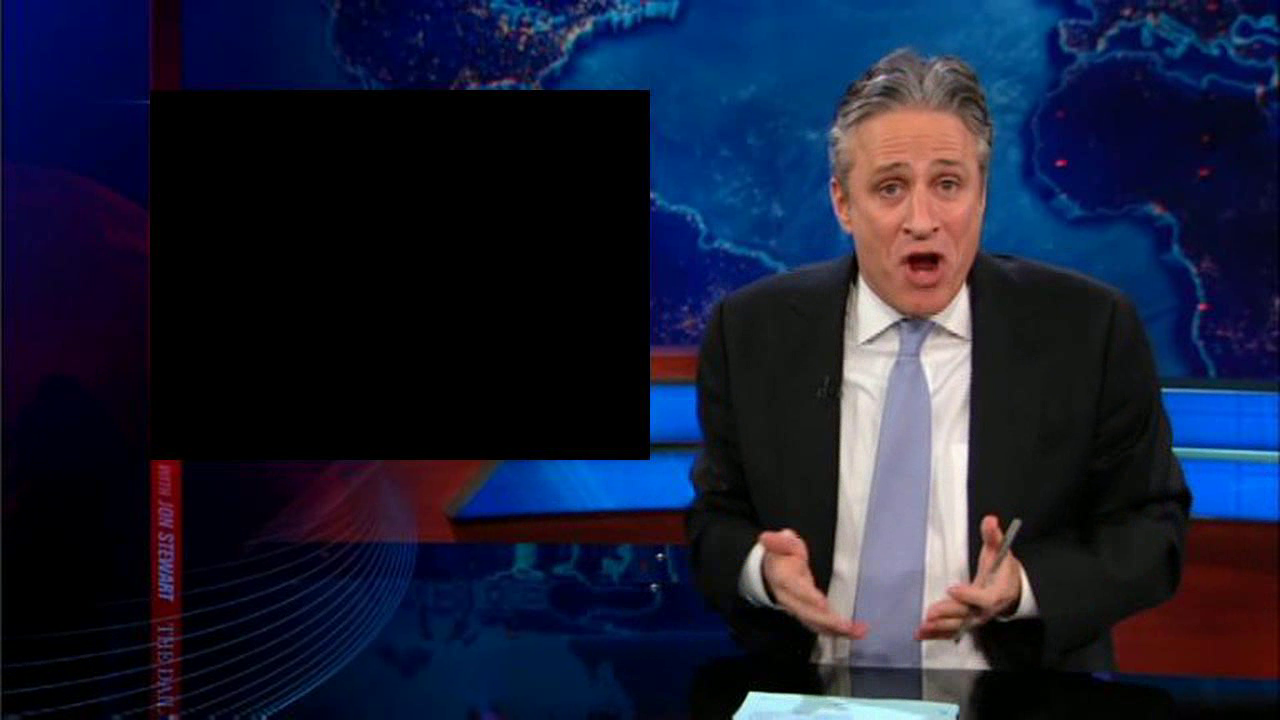}
\end{subfigure}
\begin{subfigure}{.34\textwidth}
  \centering
  \includegraphics[width=\linewidth]{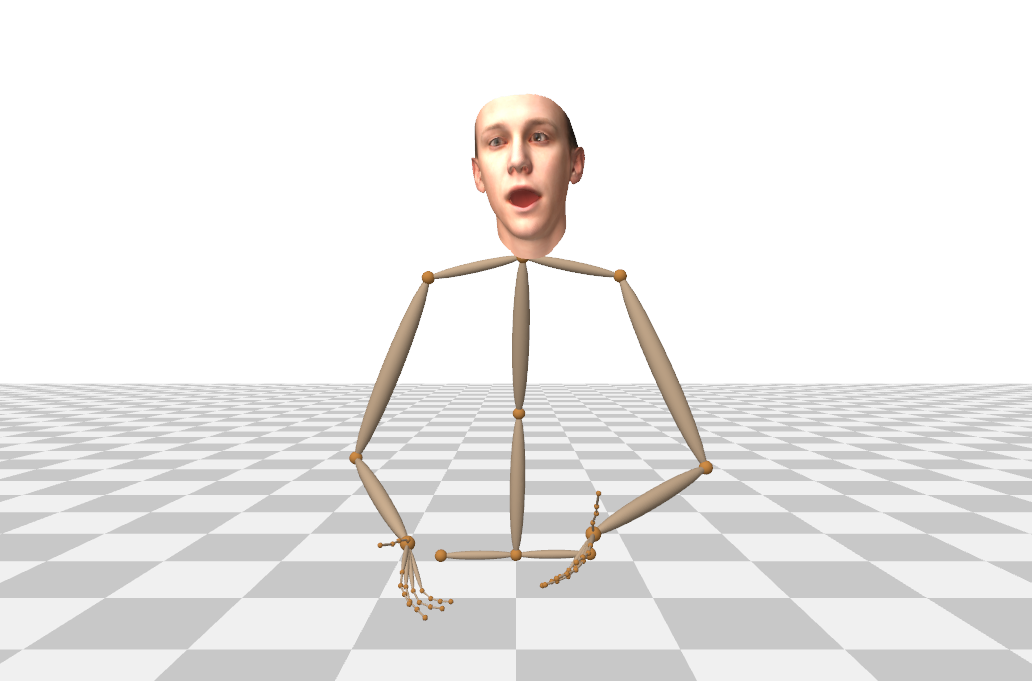}
\end{subfigure}

\caption{The 3D annotations are created from monocular in-the-wild videos using the monocular dense 3D facial reconstruction (dense face mesh visualized) approaches of Garrido \etal~\protect\cite{garrido2016reconstruction}, 3D hand pose estimation of Zhou~\etal~\protect\cite{zhou2019monocular}, and 3D body pose estimation approach of Mehta \etal~\protect\cite{mehta2019xnect}.}
    \vspace{-0.5cm}
    \label{fig:data_examples}
\end{figure*}

Figure~\ref{fig:data_examples} shows examples of the 3D facial, body and hand pose annotations extracted from in-the-wild footage.

\begin{figure*}[ht]
\centering
    
\begin{subfigure}{.40\textwidth}
  \centering
  \includegraphics[width=\linewidth]{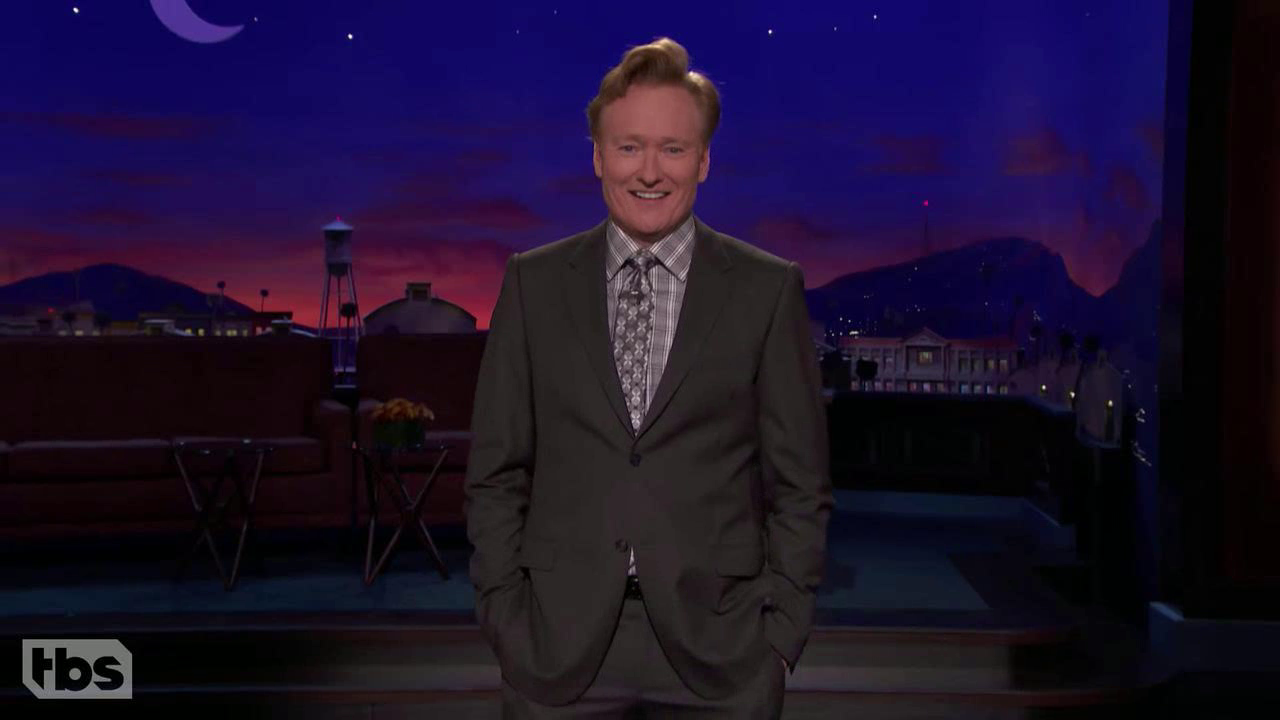}
\end{subfigure}
\begin{subfigure}{.34\textwidth}
  \centering
  \includegraphics[width=\linewidth]{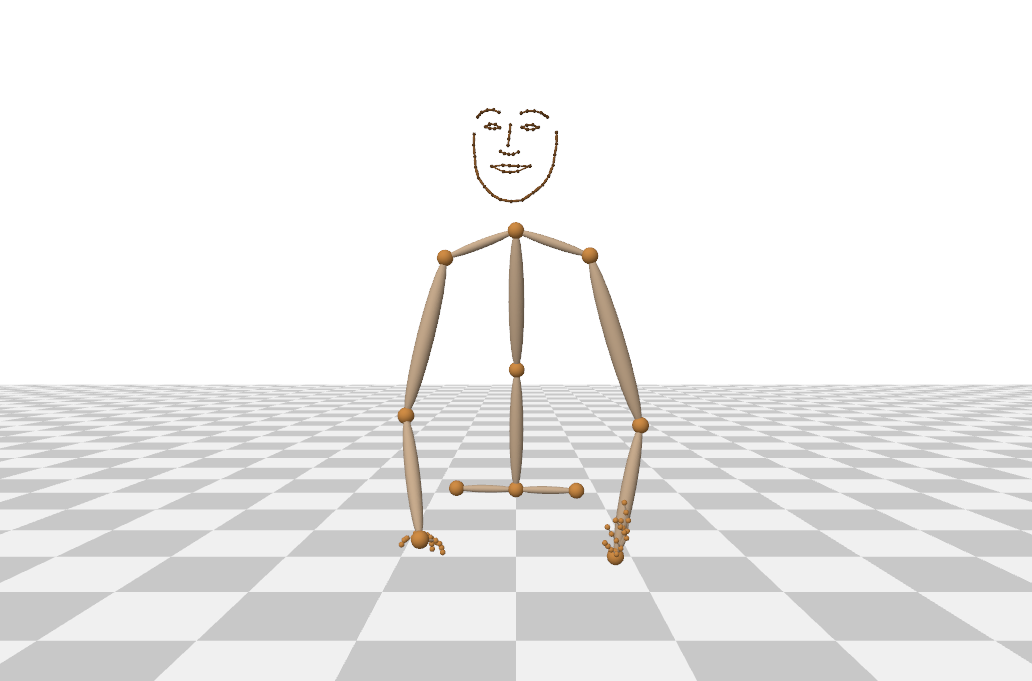}
\end{subfigure}

\vspace{1mm}

\begin{subfigure}{.40\textwidth}
  \centering
  \includegraphics[width=\linewidth]{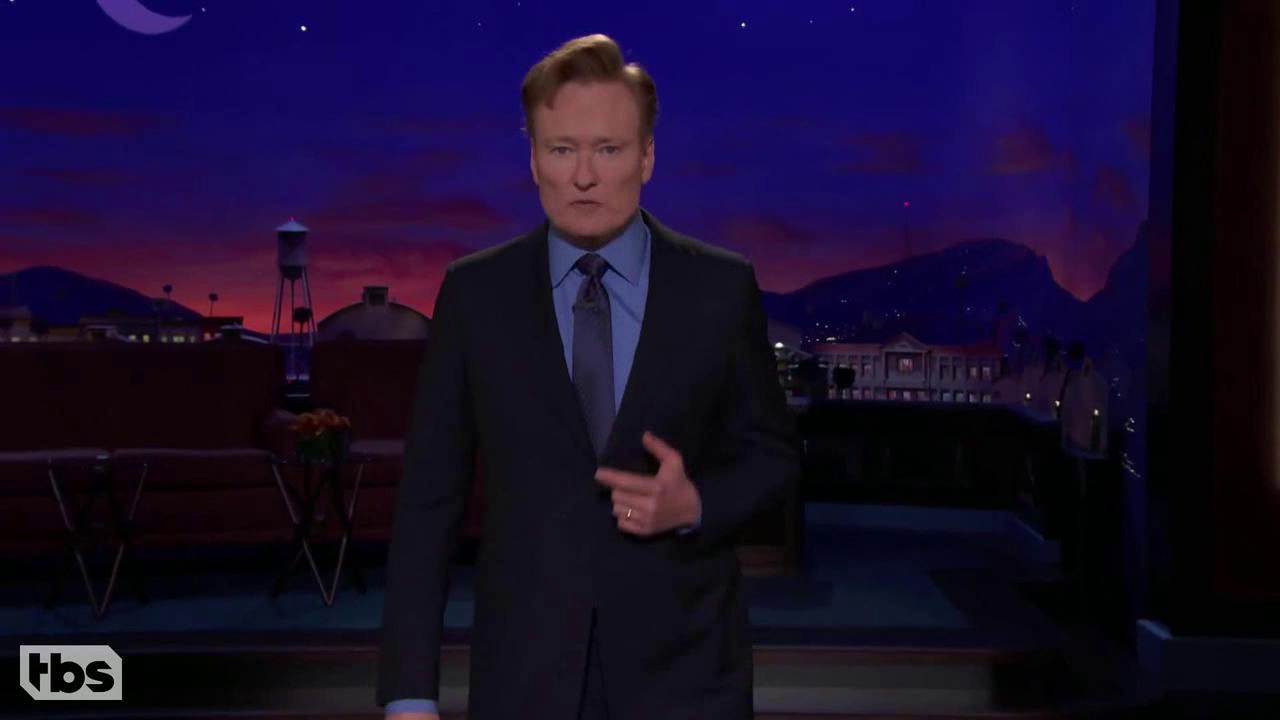}
\end{subfigure}
\begin{subfigure}{.34\textwidth}
  \centering
  \includegraphics[width=\linewidth]{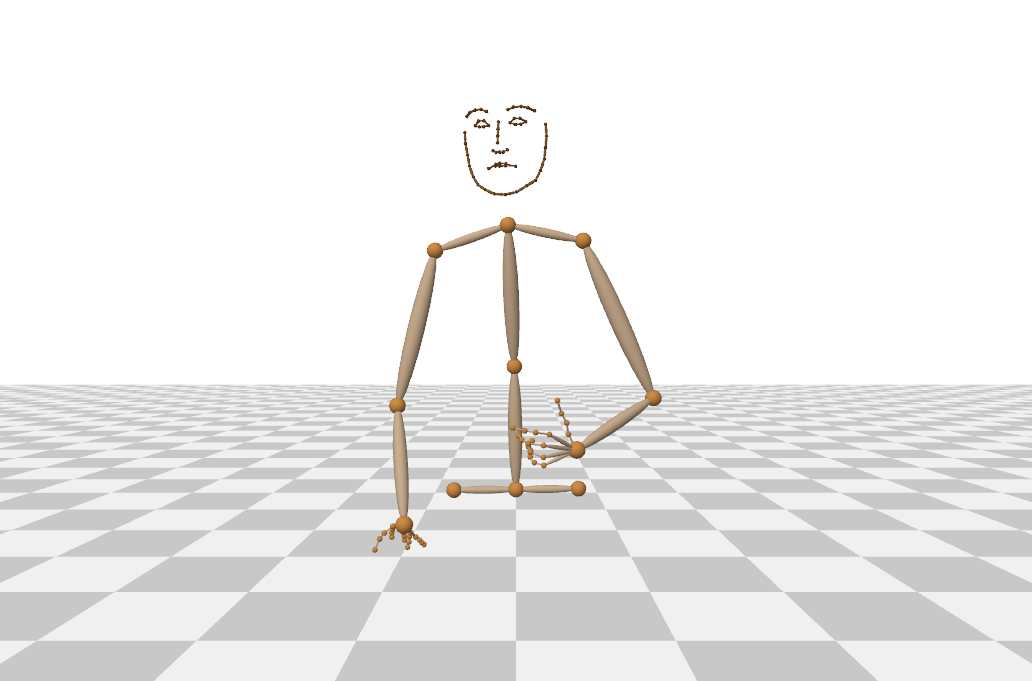}
\end{subfigure}

\vspace{1mm}

\begin{subfigure}{.40\textwidth}
  \centering
  \includegraphics[width=\linewidth]{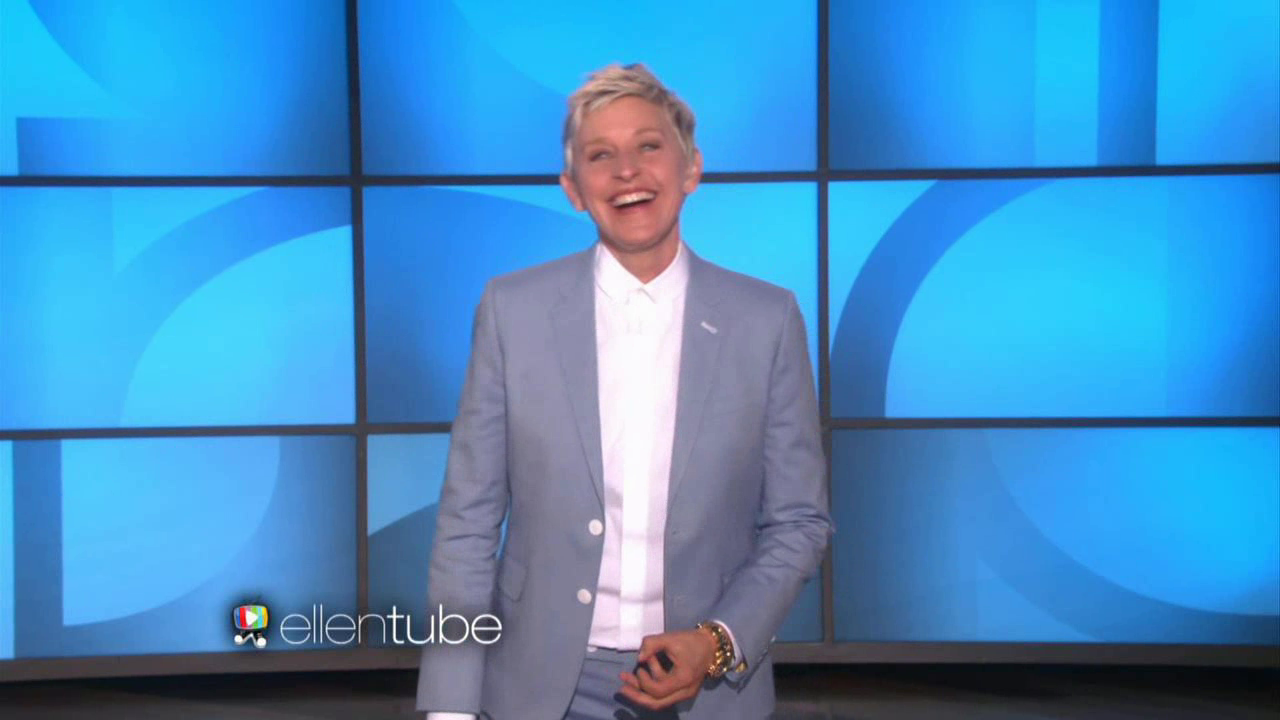}
\end{subfigure}
\begin{subfigure}{.34\textwidth}
  \centering
  \includegraphics[width=\linewidth]{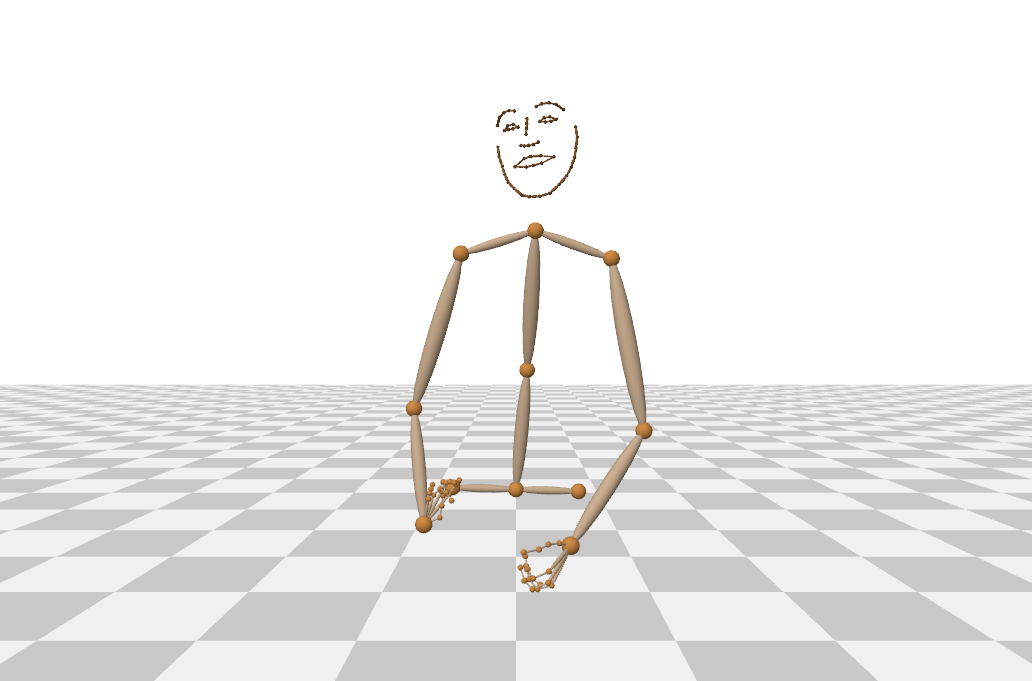}
\end{subfigure}

\vspace{1mm}

\begin{subfigure}{.40\textwidth}
  \centering
  \includegraphics[width=\linewidth]{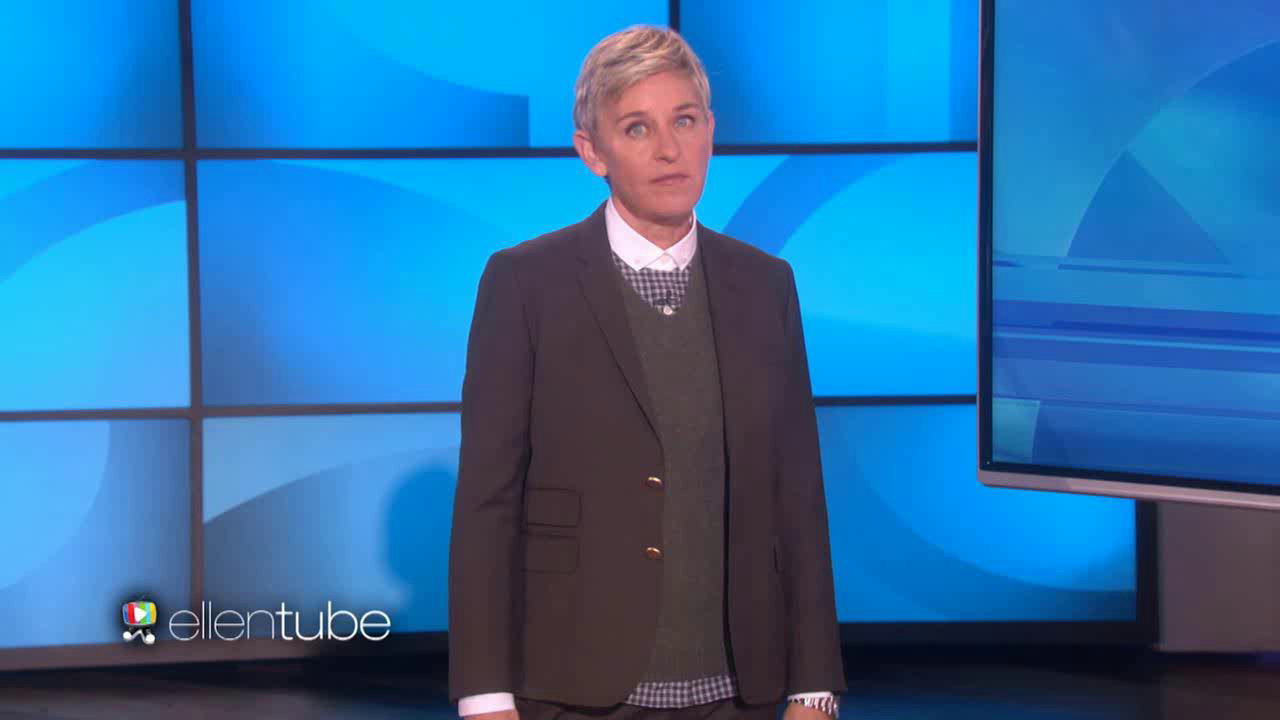}
\end{subfigure}
\begin{subfigure}{.34\textwidth}
  \centering
  \includegraphics[width=\linewidth]{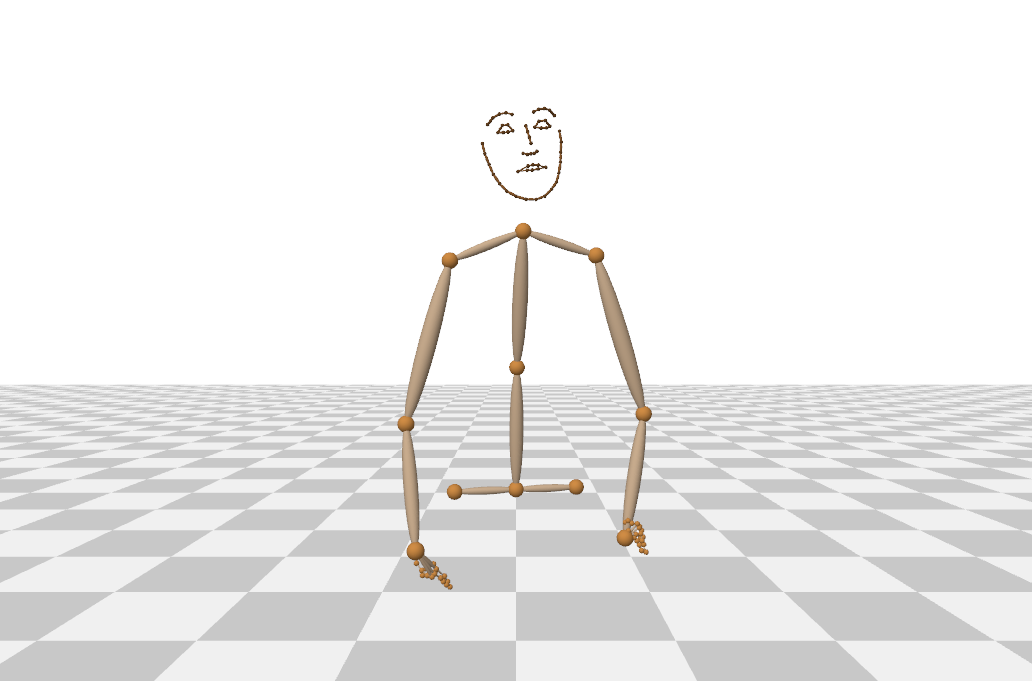}
\end{subfigure}

\vspace{1mm}

\begin{subfigure}{.40\textwidth}
  \centering
  \includegraphics[width=\linewidth]{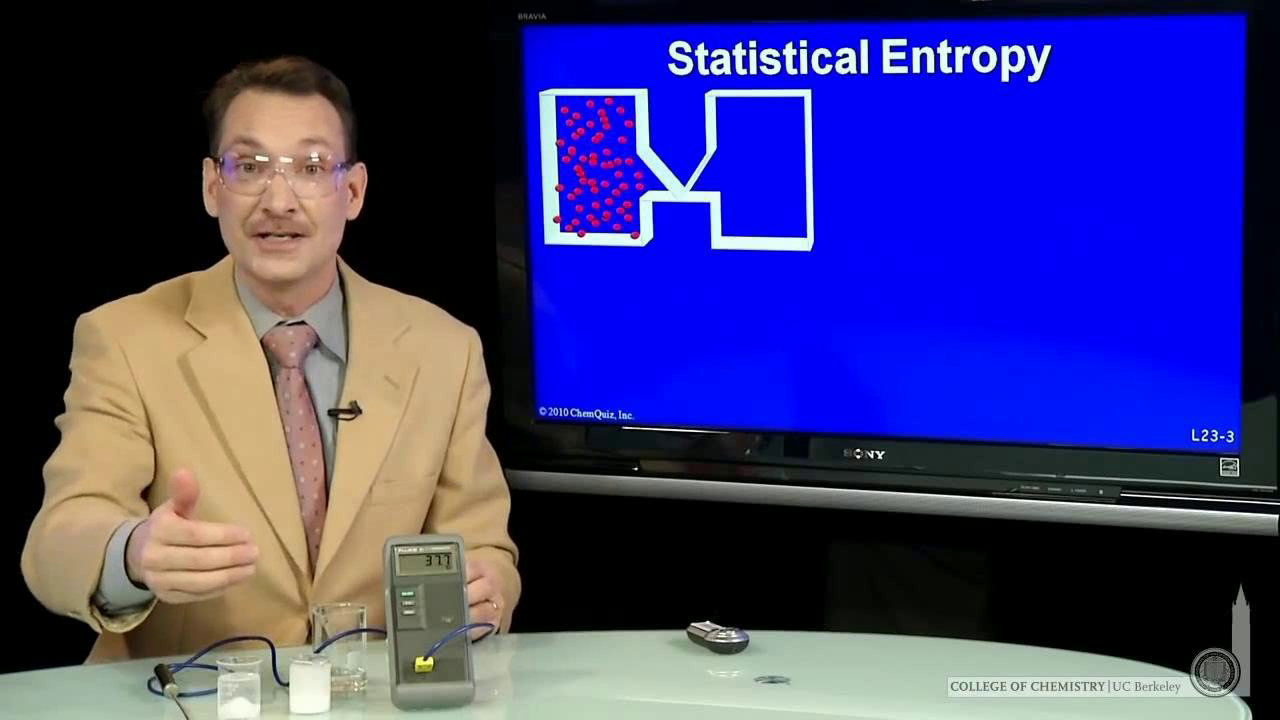}
\end{subfigure}
\begin{subfigure}{.34\textwidth}
  \centering
  \includegraphics[width=\linewidth]{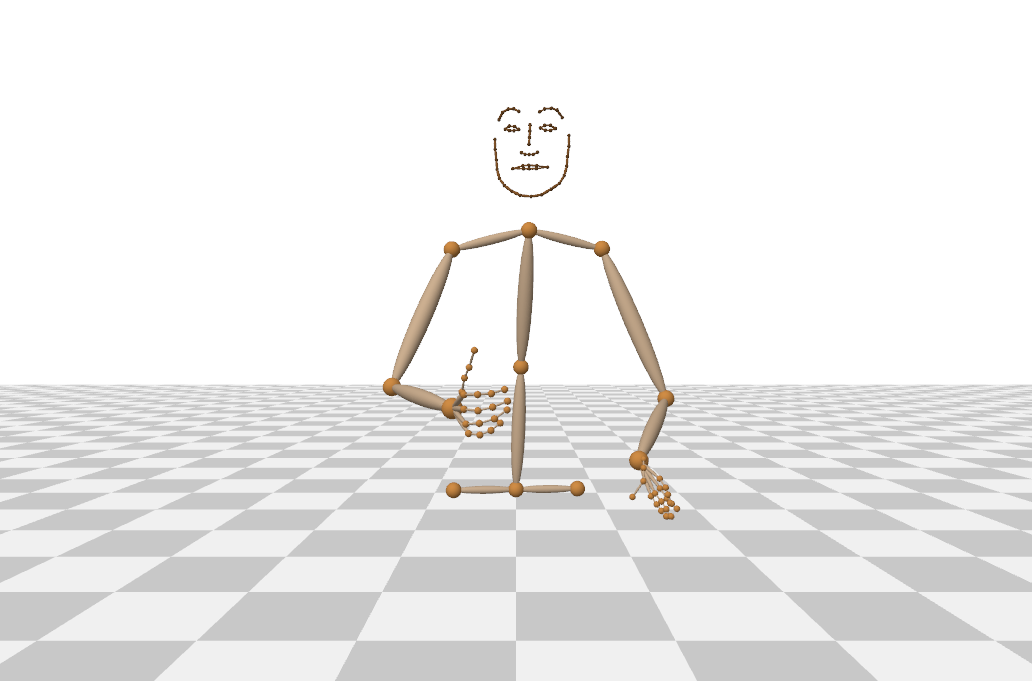}
\end{subfigure}

\vspace{1mm}

\caption{We observe that occlusion scenarios are commonly found in the video corpus, and they happen frequently for standing subjects. We apply a confidence-based filter to remove these occluded frames. To ensure that we can collect sufficient training data for each subject, we set our filtering threshold to tolerate hand occlusion cases if they occur over a short time period. 
}
    \vspace{-0.5cm}
    \label{fig:failure_examples}
\end{figure*}

\subsection{Annotation Quality and Post-processing}

We create our training data by annotating in-the-wild videos using 3D monocular tracking approaches. Naturally, the quality of our pseudo ground truth is less accurate compared to using a standard multi-view motion capture or performance capture systems. Unlike Ginosar \etal \cite{ginosar2019gestures} which evaluate their automatic 2D labels against human annotations, it is not possible for us to quantitatively measure the quality of our 3D annotations. However, due to the controlled setting of the recordings, most of the tracked videos consists of commonly observed poses often found in the training set of the monocular tracking methods we employ. Through manual inspection of the tracking results, we found that the prediction is quite reliable for our task. 

As we already mentioned in the main document, we are using the average confidence of the 2D keypoint predictions over a sequence to filter out low quality data based on a certain threshold. However, we noticed that the hands are often out of view, especially in the videos where the subjects are standing, e.g. Ellen and Conan. To prevent losing a significant amount of data, our confidence threshold is designed to allow some of the occluded hand cases to be included in the dataset if they occur over a short time window. Some of these hand occlusion examples are shown in Figure \ref{fig:failure_examples}.

Please refer to the original paper of the 3D body tracker XNect \cite{mehta2019xnect}, the hand tracker \cite{zhou2019monocular}, and the face tracker \cite{garrido2016reconstruction} for the detailed quantitative performance of their respective methods on several benchmark datasets.

To improve visual quality of our output we temporally smooth our 3D body and hand pose prediction as well as the head rotation results using a Gaussian filter with a standard deviation of $\sigma = 1.5$. We also perform the same filter for the ground truth sequences of our videos.

\subsection{2D-to-3D Lifting vs. Image-to-Body Pose}
The footage from Ginosar \etal is annotated with 2D body keypoint locations. One way to achieve 3D annotations is by running state-of-the-art 2D-to-3D lifting methods  such as Martinez~\etal~\cite{Martinez_2017_ICCV} and Pavllo \etal \cite{pavllo2019} on the provided 2D keypoints. However, multiple subjects in the dataset are either seated behind a desk, or have parts of the upper body outside the image frame. 
Occluded pelvis (by desk) cause the already ambiguous 2D-to-3D lifting approaches to have no information for correctly predicting the torso leaning, which are crucial for conveying conversational gestures. Approaches such as XNect~\cite{mehta2019xnect} are designed to be robust against partial occlusion and utilize image cues to predict the potentially missing torso.

\section{Network Architecture}
\begin{figure*}[ht]
    
\begin{subfigure}{1\textwidth}
  \centering
  \includegraphics[width=\linewidth]{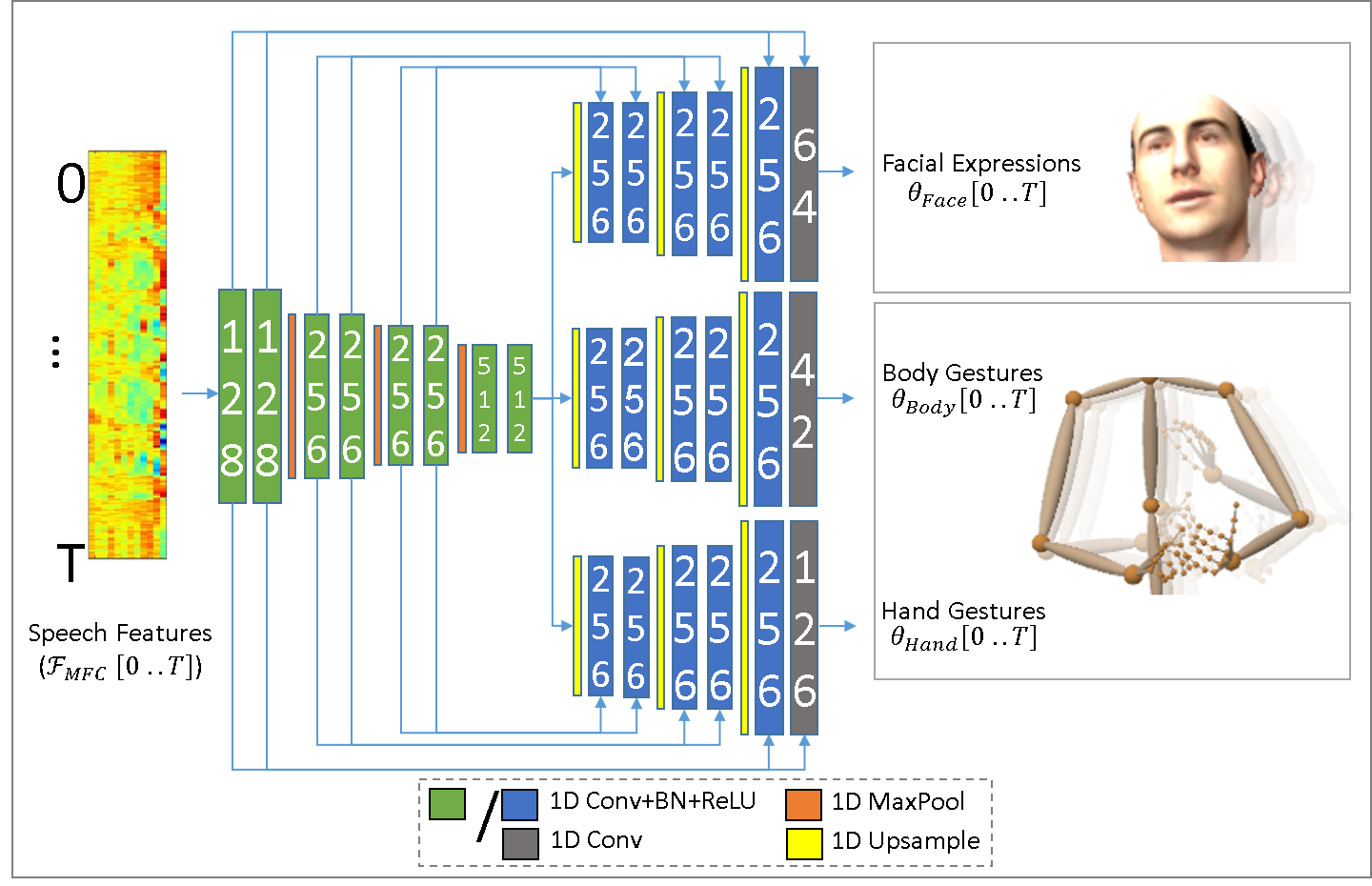}
  \caption{Generator architecture.}
\end{subfigure}
\begin{subfigure}{1\textwidth}
  \centering
  \includegraphics[width=\linewidth]{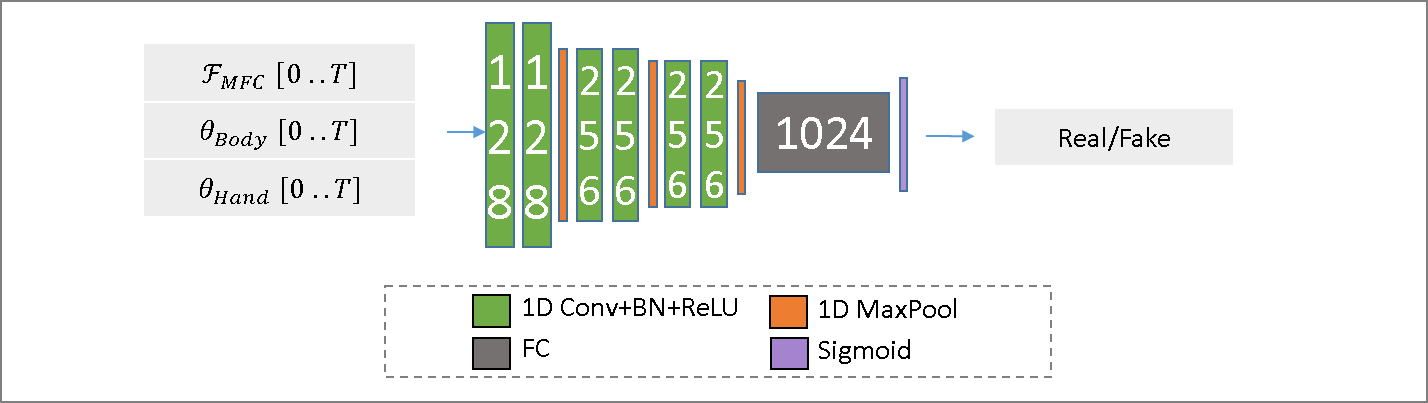}
  \caption{Discriminator architecture.}
\end{subfigure}
\caption{Details of the proposed network architecture we employ. The design of the generator has a common encoder, but separate decoders for facial expression parameters, hands, and body pose (including head pose). The numbers in the blocks represent the number of feature channels output by the block. The discriminator uses the body and hand parameters which we concatenate with the audio features as input to the network.}
\label{fig:architecture}
\end{figure*}

Details of the network architecture are described in the main document in Sec 4.1. Figure \ref{fig:architecture} shows a diagrammatic representation of the network architecture.

\begin{table}[h]
\small
    \caption{
    Quantitative result of the discriminator when trained as a audio-to-body sync/off-sync pair classifier on Oliver test sequences (higher is better).
}
    \vspace{-0.2cm}
	\label{table:sync_classifier} 
	\centering
	\begin{tabular}{c|c|c|c}
		\hline
	    Seq. length	& \makecell{In-sync pair} & \makecell{Off-sync pair} & \makecell{Combined} \\
		\hline
		\hline

	    \makecell{64 frames} & $82.7\%$ & $92.1\%$ & $87.4\%$ \\ \hline
	    \makecell{32 frames}  & $81.4\%$ & $83.1\%$ & $82.2\%$ \\ \hline
	    \makecell{16 frames}  & $74.1\%$ & $80.8\%$ & $77.5\%$ \\ \hline
	\end{tabular}
	    \vspace{-0.3cm}
\end{table} 

\section{Further Analysis of the Discriminator}

As we discussed in the main document, our proposed discriminator has the potential be extended as an independent model to score the synthesis quality of a speech-to-gesture model. One way to validate this idea is to train the model to classify whether its audio-gesture pair input is in-sync or off-sync. The ground truth audio-gesture pairs can be used directly as in-sync training examples, while off-sync training examples can be prepared by pairing the audio sequences with other random gesture sequences.
When we train our discriminator network in this setup, as shown in Table \ref{table:sync_classifier}, it can reliably classify unseen test pairs with a high accuracy. One can then argue that it is possible to extend this model as a quantitative metric for gesture synthesis methods. Unfortunately, this is currently not possible since the classifier is trained only on ground-truth motion sequences. When we tested this classifier on the baseline models, it produces inconsistent results. For example, it rates our proposed model to be more plausible than the ground truth sequences. Furthermore, it also gives the highest score to the CNN-only baseline. This assessment clearly contradicts the result of the user study. We need to create a new in-sync/off-sync dataset which contains different gesture noise characteristic if we want to extend this classifier into a more general gesture plausibility metric. 

This classifier experiment also suggests that the discriminator can provide a useful feedback signal to the generator during training by measuring the synchronization quality between audio and gesture.  A vanilla gesture-only discriminator is not capable of providing such feedback. Furthermore, this experiment also provides an insight regarding the optimal context window size, showing that using a longer training sequence leads to a better performance of the model.

{\small
\bibliographystyle{ieee_fullname}
\bibliography{article}

\begin{thebibliography}{10}\itemsep=-1pt

\bibitem{ahuja2020style}
Chaitanya Ahuja, Dong~Won Lee, Yukiko~I. Nakano, and Louis-Philippe Morency.
\newblock Style transfer for co-speech gesture animation: A multi-speaker
  conditional-mixture approach.
\newblock In Andrea Vedaldi, Horst Bischof, Thomas Brox, and Jan-Michael Frahm,
  editors, {\em ECCV}, pages 248--265. Springer International Publishing, 2020.

\bibitem{alexanderson2020style}
Simon Alexanderson, Gustav~Eje Henter, Taras Kucherenko, and Jonas Beskow.
\newblock Style-controllable speech-driven gesture synthesis using normalising
  flows.
\newblock {\em Computer Graphics Forum}, 39(2):487--496, 2020.

\bibitem{Bozkurt2013multimodal}
E. {Bozkurt}, S. {Asta}, S. {Özkul}, Y. {Yemez}, and E. {Erzin}.
\newblock Multimodal analysis of speech prosody and upper body gestures using
  hidden semi-markov models.
\newblock In {\em 2013 IEEE International Conference on Acoustics, Speech and
  Signal Processing}, 2013.

\bibitem{cao2017realtime}
Zhe Cao, Tomas Simon, Shih-En Wei, and Yaser Sheikh.
\newblock Realtime multi-person 2d pose estimation using part affinity fields.
\newblock In {\em CVPR}, 2017.

\bibitem{cassell2000embodied}
Justine Cassell.
\newblock Embodied conversational interface agents.
\newblock {\em Commun. ACM}, 43(4):70--78, Apr. 2000.

\bibitem{cassell1994animated}
Justine Cassell, Catherine Pelachaud, Norman Badler, Mark Steedman, Brett
  Achorn, Tripp Becket, Brett Douville, Scott Prevost, and Matthew Stone.
\newblock Animated conversation: Rule-based generation of facial expression,
  gesture \& spoken intonation for multiple conversational agents.
\newblock In {\em Proceedings of the 21st Annual Conference on Computer
  Graphics and Interactive Techniques}, SIGGRAPH '94, pages 413--420, New York,
  NY, USA, 1994. ACM.

\bibitem{cassell2004beat}
Justine Cassell, Hannes~H{\"o}gni Vilhj{\'a}lmsson, and Timothy Bickmore.
\newblock {\em BEAT: the Behavior Expression Animation Toolkit}, pages
  163--185.
\newblock Springer Berlin Heidelberg, Berlin, Heidelberg, 2004.

\bibitem{Cha2018TFM}
Y. {Cha}, T. {Price}, Z. {Wei}, X. {Lu}, N. {Rewkowski}, R. {Chabra}, Z. {Qin},
  H. {Kim}, Z. {Su}, Y. {Liu}, A. {Ilie}, A. {State}, Z. {Xu}, J. {Frahm}, and
  H. {Fuchs}.
\newblock Towards fully mobile {3D} face, body, and environment capture using
  only head-worn cameras.
\newblock {\em IEEE Transactions on Visualization and Computer Graphics
  (TVCG)}, 24(11):2993--3004, 2018.

\bibitem{Chiu2011how}
Chung-Cheng Chiu and Stacy Marsella.
\newblock How to train your avatar: A data driven approach to gesture
  generation.
\newblock In Hannes~H{\"o}gni Vilhj{\'a}lmsson, Stefan Kopp, Stacy Marsella,
  and Kristinn~R. Th{\'o}risson, editors, {\em Intelligent Virtual Agents},
  2011.

\bibitem{Chiu2014gesture}
Chung-Cheng Chiu and Stacy Marsella.
\newblock Gesture generation with low-dimensional embeddings.
\newblock In {\em Proceedings of the 2014 International Conference on
  Autonomous Agents and Multi-agent Systems}, AAMAS '14, 2014.

\bibitem{VOCA2019}
Daniel Cudeiro, Timo Bolkart, Cassidy Laidlaw, Anurag Ranjan, and Michael
  Black.
\newblock Capture, learning, and synthesis of {3D} speaking styles.
\newblock {\em Computer Vision and Pattern Recognition (CVPR)}, 2019.

\bibitem{Ferstl2018investigating}
Ylva Ferstl and Rachel McDonnell.
\newblock Investigating the use of recurrent motion modelling for speech
  gesture generation.
\newblock In {\em Proceedings of the 18th International Conference on
  Intelligent Virtual Agents}, IVA '18, 2018.

\bibitem{ferstl2019multi}
Ylva Ferstl, Michael Neff, and Rachel McDonnell.
\newblock Multi-objective adversarial gesture generation.
\newblock In {\em Motion, Interaction and Games}, MIG ’19, New York, NY, USA,
  2019.

\bibitem{ffmpeg}
{FFmpeg Developers}.
\newblock {FFMPEG}. ffmpeg.org, 2016.

\bibitem{garrido2016reconstruction}
Pablo Garrido, Michael Zollh{\"o}fer, Dan Casas, Levi Valgaerts, Kiran
  Varanasi, Patrick P{\'e}rez, and Christian Theobalt.
\newblock Reconstruction of personalized 3d face rigs from monocular video.
\newblock {\em ACM Transactions on Graphics (TOG)}, 35(3):28, 2016.

\bibitem{ginosar2019gestures}
S. Ginosar, A. Bar, G. Kohavi, C. Chan, A. Owens, and J. Malik.
\newblock Learning individual styles of conversational gesture.
\newblock In {\em Computer Vision and Pattern Recognition (CVPR)}. IEEE, June
  2019.

\bibitem{goldin1999role}
Susan Goldin-Meadow.
\newblock The role of gesture in communication and thinking.
\newblock {\em Trends in cognitive sciences}, 3(11):419--429, 1999.

\bibitem{Goodfellow2014generative}
Ian Goodfellow, Jean Pouget-Abadie, Mehdi Mirza, Bing Xu, David Warde-Farley,
  Sherjil Ozair, Aaron Courville, and Yoshua Bengio.
\newblock Generative adversarial nets.
\newblock In Z. Ghahramani, M. Welling, C. Cortes, N.~D. Lawrence, and K.~Q.
  Weinberger, editors, {\em Advances in Neural Information Processing Systems
  27}, pages 2672--2680. Curran Associates, Inc., 2014.

\bibitem{Haag2016bidirectional}
Kathrin Haag and Hiroshi Shimodaira.
\newblock Bidirectional lstm networks employing stacked bottleneck features for
  expressive speech-driven head motion synthesis.
\newblock In David Traum, William Swartout, Peter Khooshabeh, Stefan Kopp,
  Stefan Scherer, and Anton Leuski, editors, {\em Intelligent Virtual Agents},
  Lecture Notes in Computer Science, pages 198--207. Springer International
  Publishing, 2016.

\bibitem{deepspeech}
Awni Hannun, Carl Case, Jared Casper, Bryan Catanzaro, Greg Diamos, Erich
  Elsen, Ryan Prenger, Sanjeev Satheesh, Shubho Sengupta, Adam Coates, and
  Andrew~Y. Ng.
\newblock Deep speech: Scaling up end-to-end speech recognition, 2014.

\bibitem{hasegawa2018evaluation}
Dai Hasegawa, Naoshi Kaneko, Shinichi Shirakawa, Hiroshi Sakuta, and Kazuhiko
  Sumi.
\newblock Evaluation of speech-to-gesture generation using bi-directional lstm
  network.
\newblock In {\em Proceedings of the 18th International Conference on
  Intelligent Virtual Agents}, IVA '18, pages 79--86, New York, NY, USA, 2018.
  ACM.

\bibitem{Karras2017}
Tero Karras, Timo Aila, Samuli Laine, Antti Herva, and Jaakko Lehtinen.
\newblock Audio-driven facial animation by joint end-to-end learning of pose
  and emotion.
\newblock {\em ACM Trans. Graph.}, 36(4), July 2017.

\bibitem{kendon_2004}
Adam Kendon.
\newblock {\em Gesture: Visible Action as Utterance}.
\newblock Cambridge University Press, 2004.

\bibitem{kucherenko2019analyzing}
Taras Kucherenko, Dai Hasegawa, Gustav~Eje Henter, Naoshi Kaneko, and Hedvig
  Kjellstr\"{o}m.
\newblock Analyzing input and output representations for speech-driven gesture
  generation.
\newblock In {\em Proceedings of the 19th ACM International Conference on
  Intelligent Virtual Agents}, IVA '19, pages 97--104, New York, NY, USA, 2019.
  ACM.

\bibitem{Kucherenko2020gesticulator}
Taras Kucherenko, Patrik Jonell, Sanne van Waveren, Gustav~Eje Henter, Simon
  Alexandersson, Iolanda Leite, and Hedvig Kjellstr\"{o}m.
\newblock Gesticulator: A framework for semantically-aware speech-driven
  gesture generation.
\newblock In {\em Proceedings of the 2020 International Conference on
  Multimodal Interaction}, ICMI '20, page 242–250, New York, NY, USA, 2020.
  Association for Computing Machinery.

\bibitem{Lamere03thecmu}
Paul Lamere, Philip Kwok, Evandro Gouvêa, Bhiksha Raj, Rita Singh, William
  Walker, Manfred Warmuth, and Peter Wolf.
\newblock The {CMU SPHINX-4} speech recognition system, 2003.

\bibitem{lee2019handmotiondataset}
G. Lee, Z. Deng, S. Ma, T. Shiratori, S.S. Srinivasa, and Y. Sheikh.
\newblock Talking with hands 16.2m: A large-scale dataset of synchronized
  body-finger motion and audio for conversational motion analysis and
  synthesis.
\newblock In {\em The IEEE International Conference on Computer Vision (ICCV)},
  2019.

\bibitem{Levine2010GestureC}
Sergey Levine, Philipp Kr\"{a}henb\"{u}hl, Sebastian Thrun, and Vladlen Koltun.
\newblock Gesture controllers.
\newblock In {\em SIGGRAPH '10}, 2010.

\bibitem{Levine2009real}
Sergey Levine, Christian Theobalt, and Vladlen Koltun.
\newblock Real-time prosody-driven synthesis of body language.
\newblock {\em ACM Trans. Graph.}, 28(5):172:1--172:10, Dec. 2009.

\bibitem{flame}
Tianye Li, Timo Bolkart, Michael~J. Black, Hao Li, and Javier Romero.
\newblock Learning a model of facial shape and expression from {4D} scans.
\newblock {\em ACM Transactions on Graphics}, 36(6), Nov. 2017.
\newblock Two first authors contributed equally.

\bibitem{Liu15}
Yilong Liu, Feng Xu, Jinxiang Chai, Xin Tong, Lijuan Wang, and Qiang Huo.
\newblock Video-audio driven real-time facial animation.
\newblock {\em ACM Trans. Graph.}, 34(6):182:1--182:10, Oct. 2015.

\bibitem{mariooryad2012generating}
S. {Mariooryad} and C. {Busso}.
\newblock Generating human-like behaviors using joint, speech-driven models for
  conversational agents.
\newblock {\em IEEE Transactions on Audio, Speech, and Language Processing},
  20(8):2329--2340, Oct 2012.

\bibitem{Marsella2013virtual}
Stacy Marsella, Yuyu Xu, Margaux Lhommet, Andrew Feng, Stefan Scherer, and Ari
  Shapiro.
\newblock Virtual character performance from speech.
\newblock In {\em Proceedings of the 12th ACM SIGGRAPH/Eurographics Symposium
  on Computer Animation}, SCA '13, 2013.

\bibitem{Martinez_2017_ICCV}
Julieta Martinez, Rayat Hossain, Javier Romero, and James~J. Little.
\newblock A simple yet effective baseline for 3d human pose estimation.
\newblock In {\em The IEEE International Conference on Computer Vision (ICCV)},
  Oct 2017.

\bibitem{mcneill_2000}
David McNeill.
\newblock {\em Language and Gesture}.
\newblock Language Culture and Cognition. Cambridge University Press, 2000.

\bibitem{mehta2019xnect}
Dushyant Mehta, Oleksandr Sotnychenko, Franziska Mueller, Weipeng Xu, Mohamed
  Elgharib, Pascal Fua, Hans-Peter Seidel, Helge Rhodin, Gerard Pons-Moll, and
  Christian Theobalt.
\newblock {XNect}: Real-time multi-person {3D} motion capture with a single
  {RGB} camera.
\newblock {\em ACM Transactions on Graphics}, 39(4), July 2020.

\bibitem{Neff2008Gesture}
Michael Neff, Michael Kipp, Irene Albrecht, and Hans-Peter Seidel.
\newblock Gesture modeling and animation based on a probabilistic re-creation
  of speaker style.
\newblock {\em ACM Trans. Graph.}, 27(1):5:1--5:24, Mar. 2008.

\bibitem{pavllo2019}
Dario Pavllo, Christoph Feichtenhofer, David Grangier, and Michael Auli.
\newblock 3d human pose estimation in video with temporal convolutions and
  semi-supervised training.
\newblock In {\em Conference on Computer Vision and Pattern Recognition
  (CVPR)}, 2019.

\bibitem{Pham18}
Hai~Xuan Pham, Yuting Wang, and Vladimir Pavlovic.
\newblock End-to-end learning for 3d facial animation from speech.
\newblock In {\em Proceedings of the 20th ACM International Conference on
  Multimodal Interaction}, ICMI '18, pages 361--365, New York, NY, USA, 2018.
  ACM.

\bibitem{ffmpeg_norm}
Werner Robitza.
\newblock ffmpeg-normalize. github.com/slhck/ffmpeg-normalize, 2019.

\bibitem{ronneberger2015u}
Olaf Ronneberger, Philipp Fischer, and Thomas Brox.
\newblock U-net: Convolutional networks for biomedical image segmentation.
\newblock In {\em International Conference on Medical image computing and
  computer-assisted intervention}, pages 234--241. Springer, 2015.

\bibitem{Sadoughi2018novel}
N. {Sadoughi} and C. {Busso}.
\newblock Novel realizations of speech-driven head movements with generative
  adversarial networks.
\newblock In {\em 2018 IEEE International Conference on Acoustics, Speech and
  Signal Processing (ICASSP)}, 2018.

\bibitem{Sadoughi2017speech}
Najmeh Sadoughi and Carlos Busso.
\newblock Speech-driven animation with meaningful behaviors.
\newblock {\em Speech Communication}, 110:90 -- 100, 2019.

\bibitem{Sadoughi2014speech}
Najmeh Sadoughi, Yang Liu, and Carlos Busso.
\newblock Speech-driven animation constrained by appropriate discourse
  functions.
\newblock In {\em Proceedings of the 16th International Conference on
  Multimodal Interaction}, ICMI '14, pages 148--155, New York, NY, USA, 2014.
  ACM.

\bibitem{Sadoughi2017meaningful}
Najmeh Sadoughi, Yang Liu, and Carlos Busso.
\newblock Meaningful head movements driven by emotional synthetic speech.
\newblock {\em Speech Communication}, 95:87 -- 99, 2017.

\bibitem{salimans2016improved}
Tim Salimans, Ian Goodfellow, Wojciech Zaremba, Vicki Cheung, Alec Radford, Xi
  Chen, and Xi Chen.
\newblock Improved techniques for training gans.
\newblock In D.~D. Lee, M. Sugiyama, U.~V. Luxburg, I. Guyon, and R. Garnett,
  editors, {\em Advances in Neural Information Processing Systems 29}, pages
  2234--2242. Curran Associates, Inc., 2016.

\bibitem{Saragih2011}
Jason~M. Saragih, Simon Lucey, and Jeffrey~F. Cohn.
\newblock Deformable model fitting by regularized landmark mean-shift.
\newblock {\em IJCV}, 91(2), 2011.

\bibitem{shlizerman2018audio}
Eli Shlizerman, Lucio Dery, Hayden Schoen, and Ira Kemelmacher-Shlizerman.
\newblock Audio to body dynamics.
\newblock In {\em Proceedings of the IEEE Conference on Computer Vision and
  Pattern Recognition}, pages 7574--7583, 2018.

\bibitem{suwajanakorn2017obama}
Supasorn Suwajanakorn, Steven~M. Seitz, and Ira Kemelmacher-Shlizerman.
\newblock Synthesizing obama: Learning lip sync from audio.
\newblock {\em ACM Trans. Graph.}, 2017.

\bibitem{takeuchi2017speech}
Kenta Takeuchi, Dai Hasegawa, Shinichi Shirakawa, Naoshi Kaneko, Hiroshi
  Sakuta, and Kazuhiko Sumi.
\newblock Speech-to-gesture generation: A challenge in deep learning approach
  with bi-directional lstm.
\newblock In {\em Proceedings of the 5th International Conference on Human
  Agent Interaction}, HAI '17, pages 365--369, New York, NY, USA, 2017. ACM.

\bibitem{takeuchi2017creating}
Kenta Takeuchi, Souichirou Kubota, Keisuke Suzuki, Dai Hasegawa, and Hiroshi
  Sakuta.
\newblock Creating a gesture-speech dataset for speech-based automatic gesture
  generation.
\newblock In Constantine Stephanidis, editor, {\em HCI International 2017 --
  Posters' Extended Abstracts}, pages 198--202, Cham, 2017. Springer
  International Publishing.

\bibitem{Taylor2017}
Sarah Taylor, Taehwan Kim, Yisong Yue, Moshe Mahler, James Krahe,
  Anastasio~Garcia Rodriguez, Jessica Hodgins, and Iain Matthews.
\newblock A deep learning approach for generalized speech animation.
\newblock {\em ACM Trans. Graph.}, 36(4), July 2017.

\bibitem{Tzirakis19}
Panagiotis Tzirakis, Athanasios Papaioannou, Alexander Lattas, Michail
  Tarasiou, Bj{\"{o}}rn~W. Schuller, and Stefanos Zafeiriou.
\newblock Synthesising 3d facial motion from "in-the-wild" speech.
\newblock {\em CoRR}, abs/1904.07002, 2019.

\bibitem{pytorch-unet}
Naoto Usuyama.
\newblock github.com/usuyama/pytorch-unet, 2018.

\bibitem{van1998persona}
Susanne Van~Mulken, Elisabeth Andr{\'e}, and Jochen M{\"u}ller.
\newblock The persona effect: How substantial is it?
\newblock In {\em People and computers XIII}, pages 53--66. Springer, 1998.

\bibitem{Yoon2020Trimodal}
Youngwoo Yoon, Bok Cha, Joo-Haeng Lee, Minsu Jang, Jaeyeon Lee, Jaehong Kim,
  and Geehyuk Lee.
\newblock Speech gesture generation from the trimodal context of text, audio,
  and speaker identity.
\newblock {\em ACM Transactions on Graphics}, 39(6), 2020.

\bibitem{Yoon2018robots}
Y. {Yoon}, W. {Ko}, M. {Jang}, J. {Lee}, J. {Kim}, and G. {Lee}.
\newblock Robots learn social skills: End-to-end learning of co-speech gesture
  generation for humanoid robots.
\newblock In {\em 2019 International Conference on Robotics and Automation
  (ICRA)}, pages 4303--4309, 2019.

\bibitem{zhou2019monocular}
Yuxiao Zhou, Marc Habermann, Weipeng Xu, Ikhsanul Habibie, Christian Theobalt,
  and Feng Xu.
\newblock Monocular real-time hand shape and motion capture using multi-modal
  data.
\newblock In {\em {IEEE} Conference on Computer Vision and Pattern Recognition
  (CVPR)}, pages 0--0, 2020.

\end{thebibliography}
}

\end{document}